\documentclass[lettersize,journal]{IEEEtran}
\usepackage{amsmath,amsfonts}
\usepackage{algorithmic}
\usepackage{algorithm}
\usepackage{array}
\usepackage[caption=false,font=normalsize,labelfont=sf,textfont=sf]{subfig}
\usepackage{textcomp}
\usepackage{stfloats}
\usepackage{url}
\usepackage{verbatim}
\usepackage{graphicx}
\usepackage{cite}
\usepackage{xcolor}
\usepackage{tabularx,ragged2e}
\usepackage{booktabs}
\usepackage{makecell}   
\usepackage{multirow}
\usepackage{flushend}
\usepackage{caption}
\usepackage{balance}
\begin{document}

\title{De-Simplifying Pseudo Labels to Enhancing Domain Adaptive Object Detection}

        % <-this % stops a space
\author{Zehua~Fu,
        Chenguang~Liu,
        Yuyu~Chen,
        Jiaqi~Zhou,
        % Zhenghui~Hu,
        Qingjie~Liu*,~\IEEEmembership{Member,~IEEE},
        Yunhong~Wang,~\IEEEmembership{Fellow,~IEEE}
        % <-this % stops a space
        % <-this % stops a space
\thanks{

* Corresponding author. 
% \\ \indent
Zehua Fu is with Hangzhou Innovation Institute, Beihang University, Hangzhou 310051, China.
Qingjie Liu and Yunhong Wang are with the State Key Laboratory of Virtual Reality Technology and Systems, Beihang University, Beijing 100191, China, and also with Hangzhou Innovation Institute, Beihang University, Hangzhou 310051, China. 
Chenguang Liu, Yuyu Chen and Jiaqi Zhou are with the State Key Laboratory of Virtual Reality Technology and Systems, Beihang University, Beijing 100191, China. 
Email:~zuowenhang@gmail.com,~\{zehua\_fu, liuchenguang, yuyu\_chen, gracciechou, qingjie.liu, yhwang\}@buaa.edu.cn.
}
}% <-this % stops a space

\markboth{Journal of \LaTeX\ Class Files,~Vol.~14, No.~8, August~2021}%
{Shell \MakeLowercase{\textit{et al.}}: A Sample Article Using IEEEtran.cls for IEEE Journals}

\maketitle

\begin{abstract}
Despite its significant success, object detection in traffic and transportation scenarios requires time-consuming and laborious efforts in acquiring high-quality labeled data. Therefore, Unsupervised Domain Adaptation (UDA) for object detection has recently gained increasing research attention.
UDA for object detection has been dominated by domain alignment methods, which achieve top performance. 
Recently, self-labeling methods have gained popularity due to their simplicity and efficiency. 
In this paper, we investigate the limitations that prevent self-labeling detectors from achieving commensurate performance with domain alignment methods.
Specifically, we identify the high proportion of simple samples during training, i.e., the simple-label bias, as the central cause. 
We propose a novel approach called De-Simplifying Pseudo Labels (DeSimPL) to mitigate the issue. 
DeSimPL utilizes an instance-level memory bank to implement an innovative pseudo label updating strategy. 
Then, adversarial samples are introduced during training to enhance the proportion. 
Furthermore, we propose an adaptive weighted loss to avoid the model suffering from an abundance of false positive pseudo labels in the late training period.
Experimental results demonstrate that DeSimPL effectively reduces the proportion of simple samples during training, leading to a significant performance improvement for self-labeling detectors. 
Extensive experiments conducted on four benchmarks validate our analysis and conclusions.
\end{abstract}


\section{Introduction}
Object detection~\cite{ren2015faster, redmon2016you, liu2016ssd, Crawford_Pineau_2019, Li_Huang_Hua_Zhang_2021, Hou_Lu_Xue_Li_2022, He_Dong_Lin_Lau_2023, Fengd22, ZhangLLW22, Kimwj22, Luyf23, Jian24} with deep learning is a pivotal component of computer vision, widely applied in transportation for tasks such as traffic monitoring, autonomous driving, and parking assistance. In autonomous driving, robust object detection is essential for tasks like vehicle and pedestrian recognition, which directly impact safety and navigation. However, the performance of these models often relies on large volumes of annotated data, which are costly to collect and challenging to acquire in diverse real-world scenarios, such as varying weather conditions, geographic locations, or traffic densities.
To address this issue, Unsupervised Domain Adaptation (UDA) methods have been developed to enable models to adapt to new domains without requiring additional annotations. UDA is particularly critical in autonomous driving, where domain gaps—such as those between simulation data and real-world environments or between datasets from different cities—frequently occur. By improving cross-domain adaptability, UDA methods enhance the robustness and reliability of object detection models, ensuring their applicability across diverse traffic scenarios.

\begin{figure}[t]
   \begin{center}
   \includegraphics[width=.95\linewidth]{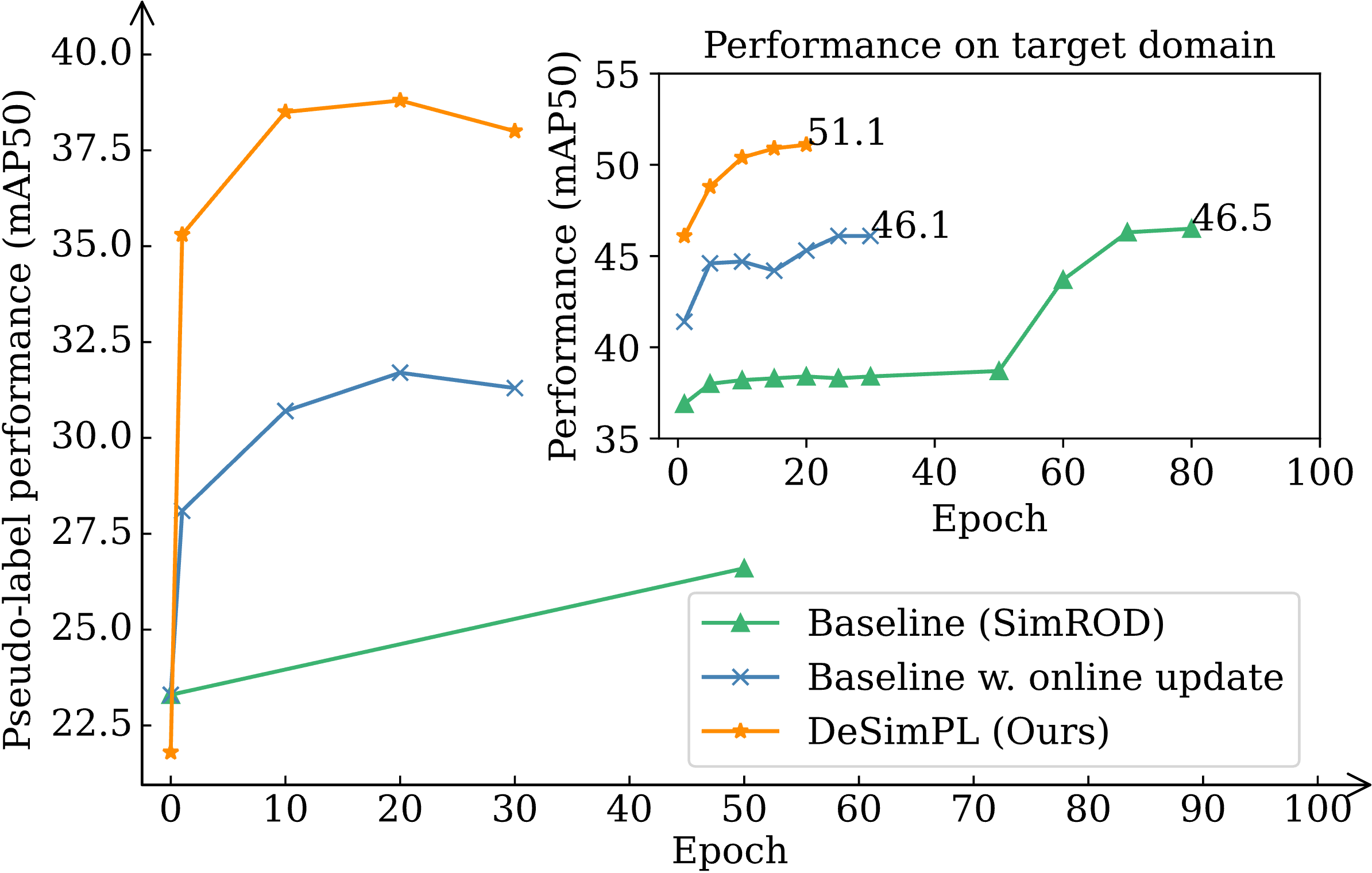}
   \end{center}
   \caption{
   In the self-labeling paradigm, online updating is an effective way to enhance the pseudo label performance and enable the model to converge quickly. 
   Nevertheless, high-quality pseudo labels do not necessarily improve the model performance on the target domain (46.1\% vs 46.5\%, w/w.o online update, VOC $\rightarrow$ Comic). 
   We identify the reason for this is the simple-label bias and introduce DeSimPL as a solution. 
   DeSimPL improves the baseline by a large margin (51.1\% vs 46.5\%).
   }
   \label{fig:simple_samples_increase}
\end{figure}

UDA in object detection has been dominated by domain alignment methods \cite{chen2018domain, saito2019strong, hsu2020every, sindagi2020prior, vs2021mega, he2019multi, hsu2020progressive, kim2019diversify, zheng2020cross, zhu2019adapting, jiangdecoupled, li2022sigma}, which utilize adversarial training with a domain discriminator and detector to learn domain-invariant features.
While consistently achieving state-of-the-art results, domain alignment methods necessitate non-trivial architecture modifications, such as gradient reversal layers, domain classifiers, or specialized modules \cite{ramamonjison2021simrod}. 
Recently, self-labeling methods~\cite{roychowdhury2019automatic, khodabandeh2019robust, kim2019self, zhao2020collaborative, yu2019unsupervised, munir2021sstn, wang2021robust, ramamonjison2021simrod, inoue2018cross} have gained popularity due to their simplicity and efficiency.
These methods utilize highly confident target predictions of the source-trained detector model, i.e., pseudo labels, to iteratively improve the target detector.
\par
\begin{figure*}[t!]
   \begin{center}
   \includegraphics[width=\linewidth]{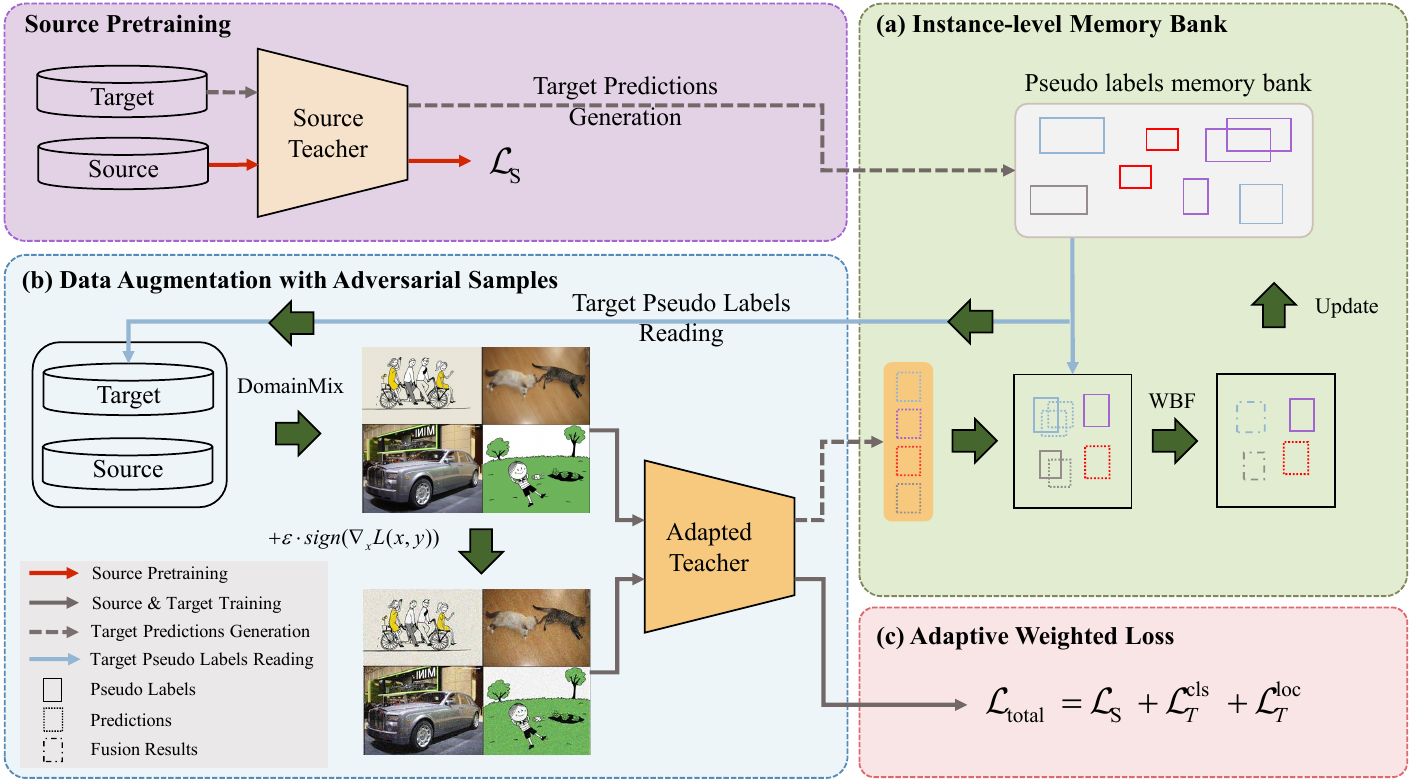}
   \end{center}
      \caption{
      % \textcolor{red}{
      Our DeSimPL comprises three components: an online update pseudo label strategy based on the instance-level memory bank, a data augmentation strategy combined with adversarial examples, and an adaptive weighting algorithm based on the pseudo-label localization loss.
      % }
      }
   \label{fig:overall_method}
\end{figure*}
While these methods are simple and efficient, most of them are inferior to domain alignment methods.
An inquiry that naturally emerges is whether a self-labeling detector can achieve comparable performance to domain alignment methods.
In this paper, we introduce a self-labeling UDA object detector that achieves consistent and comparable performance to domain alignment methods, such as D-adapt  \cite{jiangdecoupled} and SIGMA++  \cite{li2023sigma++}. 
To attain this outcome, the large scale of the simple samples during training, namely simple-label bias (as shown in Figure~\ref{fig:simple_samples_increase}), is identified as the main hindrance preventing self-labeling methods from achieving state-of-the-art accuracy.
\par
\begin{figure}[tb]
   \begin{center}
   \includegraphics[width=.95\linewidth]{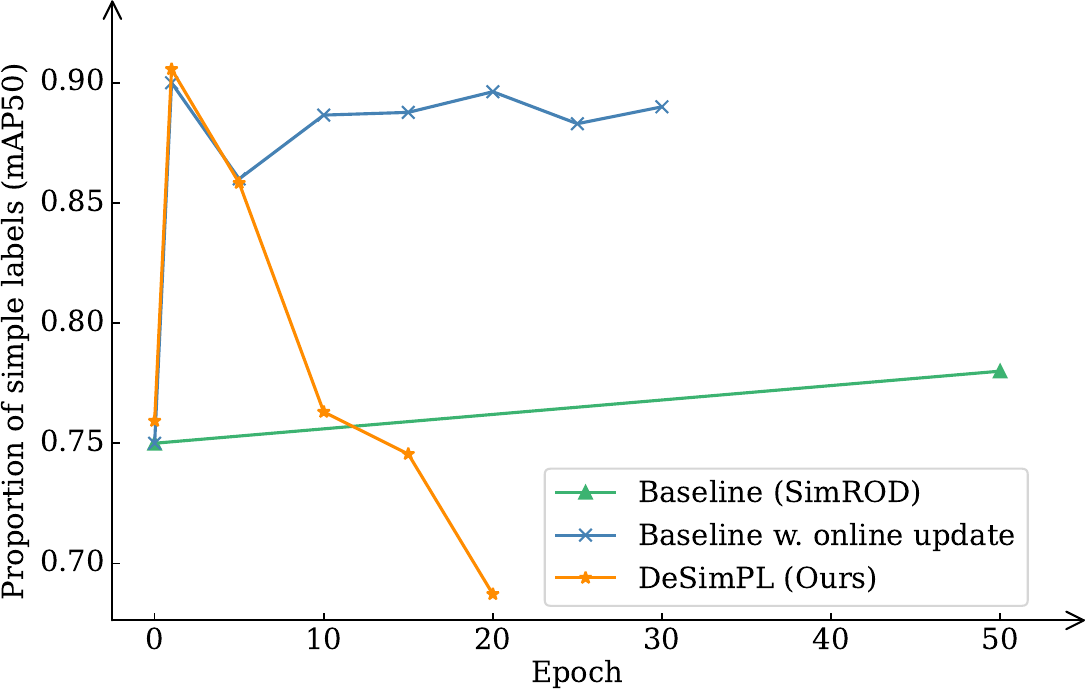}
   \end{center}
   \caption{
      The proportion variations of simple samples (i.e., samples with a loss value $\le$ 0.3) with true positive (TP) pseudo labels of different methods.
   For baseline w. online update method, the proportion of simple samples remains at a relatively high level as the model iteratively updates pseudo labels.
   After applying our DeSimPL, the proportion of simple samples gradually decreases as the pseudo labels update, enabling the model to attain the highest performance on the target domain with the fastest convergence speed.
   }
   \label{fig:pslabel_performance}
\end{figure}

To address the aforementioned limitations, we propose De-Simplifying Pseudo Labels (DeSimPL), an approach that alters the pseudo labels during training to diminish simple samples, thus bolstering the model's performance on the target domain. 
To realize this objective, we first establish and maintain a dynamically updating instance-level memory bank to store historical pseudo labels. 
In particular, this memory bank is periodically updated using the weighted box fusion strategy whenever the latest pseudo labels are generated, thereby preventing the pseudo label from overfitting.
Subsequently, we incorporate adversarial noise into the training process to boost the number of hard samples and adaptively adjust the loss of the target domain with an adaptive weighted loss to further enhance the model's performance.
The experimental results in Figure~\ref{fig:pslabel_performance} 
demonstrate that adopting the proposed DeSimPL can effectively alleviates the simple-label bias and significantly improve the performance of domain adaptive object detection.
\par
The main contributions of this paper are summarized as follows.
First, we identify a critical problem in the self-labeling methodology that limits the model's performance as training progresses, called simple-label bias. 
Second, we propose a simple yet effective method named DeSimPL to alleviate the simple-label bias.
The core of the method is a new pseudo label update strategy that consists of three main components including an instance-level memory bank, adversarial data augmentations, and an adaptive weighted loss. 
Third, we demonstrate the effectiveness of our method through extensive experiments and achieve state-of-the-art results in four domain adaptive object detection benchmarks.

\section{Related Work}
Domain adaptive object detection plays a crucial role in traffic-related scenarios. 
In this section, we present a comprehensive review of domain adaptive object detection.
Various domain adaptation methods have been proposed to address the problem of domain shift~\cite{ZhangLLW22, LiJQ22, Liuhj24, Wangxw24}. These methods can be broadly classified into two categories, Domain-alignment based methods and self-labeling based methods. 
\par
Domain-alignment is the mainstream paradigm for domain adaptive object detection, which utilizes a domain discriminator to align the features at different levels.
DA-Faster \cite{chen2018domain} was the first work to address UDA based on the Faster-RCNN \cite{ren2015faster} for global and instance-level feature alignment.
Saito et al. \cite{saito2019strong} used focal loss \cite{lin2017focal} instead of cross-entropy loss for global alignment, focusing feature alignment more on the foreground. 
Zheng et al. \cite{zheng2020cross} added an attention module to assist in alignment, further improving the foreground alignment. 
Rezaeianaran et al. \cite{rezaeianaran2021seeking} and Zhu et al. \cite{zhu2019adapting} clustered the features of proposals before feature alignment. 
Some other methods such as CRDA \cite{xu2020exploring} and MCAR \cite{zhao2020adaptive} use classification as auxiliary tasks for feature alignment. 
D-adapt \cite{jiangdecoupled} decouples adversarial adaptation and detector training to further enhance performance. 
However, the challenge is to determine where to add the alignment module and discriminator in the model, and these modules require additional training.
Another approach to UDA is based on style transfer using Generative Adversarial Networks (GANs) to convert source domain images into target domain style images or vice versa, in order to reduce the domain gap and improve detector performance in the target domain \cite{hsu2020progressive, zhang2019cycle, chen2020harmonizing, rodriguez2019domain, hoffman2018cycada}. 
Additionally, there are methods that aim to diversify the image styles during training \cite{kim2019diversify, rodriguez2019domain}, ensuring that the detector is not biased towards any particular style. However, these methods require pre-training of a style transfer model and additional training time, making them computationally expensive.
\par
Recently, self-labeling has emerged as a promising alternative, gaining momentum in the research community.
Self-labeling techniques generate pseudo-labels for target domain data using a detector trained on the source domain. These pseudo-labels are utilized to retrain the model on the target domain, with a major focus being reducing the noise of the labels. Various methods have been proposed to accomplish this. For example, Roychowdhry et al. \cite{roychowdhury2019automatic} proposed a self-labeling approach for single-class object detection that utilizes video data in the target domain to automatically generate pseudo-labels. Khodabandeh et al. \cite{khodabandeh2019robust} use additional classifiers to denoise the pseudo-labels by refining the category of each pseudo-label in the target domain using an image classifier pre-trained on large-scale data. Meanwhile, Zhao et al. \cite{zhao2020collaborative} introduced a domain adaptation method based on the co-training of RPN and head classification network. The method utilizes the high-confidence output of one of the networks to train the other. SimROD \cite{ramamonjison2021simrod} is a self-labeling approach that utilizes a teacher model to direct the student model, drawing on the experience of classic semi-supervised methods like STAC \cite{sohn2020simple} and SoftTeacher \cite{zhou2021instant}. However, the distinguishing feature of SimROD is that it creates pseudo-labels for the target domain using a large-scale teacher model and updates the pseudo-labels once while training the teacher model. Consequently, the highly accurate pseudo-labels generated by the teacher model are used to supervise the training of the student model. SimROD posits that the use of pseudo-labels produced by large teacher models can significantly improve the performance of the student model since larger models are believed to be more robust to domain shift.
\par
The SoftTeacher approach\cite{zhou2021instant} has shown the efficacy of updating pseudo-labels in semi-supervised learning. However, in the context of domain adaptation, continually updating pseudo-labels during training can result in a higher proportion of simpler samples, thereby impeding the detection performance. To address this issue and boost the model's detection performance, we suggest updating the pseudo-labels during the training phase of the teacher model in SimROD and enhancing the updating approach.

\section{Methodology}

\subsection{\textbf{Problem statement}}
In the unsupervised domain adaptation (UDA), we have a labeled source domain dataset $D_{s}=\{(x_{i},y_{i})\}$, where $x_{i}$ is an image and $y_{i}$ is its corresponding labeling information, including the category and coordinates of the objects in the images.
Similarly, we denote the unlabeled target domain dataset as $D_{t}=\{(x_{j})\}$. 
Among them, there is a domain shift between the source domain and the target domain, namely $p_S(y|x) = p_T(y|x)$ but $p_S(x)\neq p_T(x)$. 
\par
As the target domain lacks labeled data, the noisy initial pseudo labels generated by the source model pose significant challenges for object detection on the target domain. 
Enhancing the quality of pseudo labels during the model adaptation process is a key strategy for improving the overall performance of UDA models. In this paper, we propose a simple yet effective method for improving the quality of pseudo labels during model adaptation.

\subsection{Simple-label bias}
Intuitively, updating the pseudo labels to have a better quality is a straightforward way to improve the model performance. 
Inspired by~\cite{yang2021st3d}, we conduct experiments on a typical self-labeling method (e.g., SimROD) by improving it with fixed and shortened pseudo-label update intervals (i.e., SimROD w. online update).  
As depicted in Figure~\ref{fig:simple_samples_increase}, online updating notably enhances the quality of pseudo labels; however, trained with the labels, the model's performance on target domain remains a similar overall performance to that of SimROD (SimROD: 46.5\% mAP; SimROD w. online update: 46.1\% mAP).
To figure out the reason behind it, we investigate how the samples with pseudo labels contribute to the training, typically focusing on the samples with a loss less than 0.3, i.e., simple samples, since these samples have small contributions. We visualize them in Figure 3 and observe 
that these simple samples maintain a high proportion as the model iteratively updates pseudo labels.
That is to say, as the pseudo labels improve, they do not contribute the model learning. We call this the simple-label bias and believe it is a key cause of impending the capacity of the model to achieve further improvements in performance during training.

\begin{figure}[t]
    \centering
    \includegraphics[width=\linewidth]{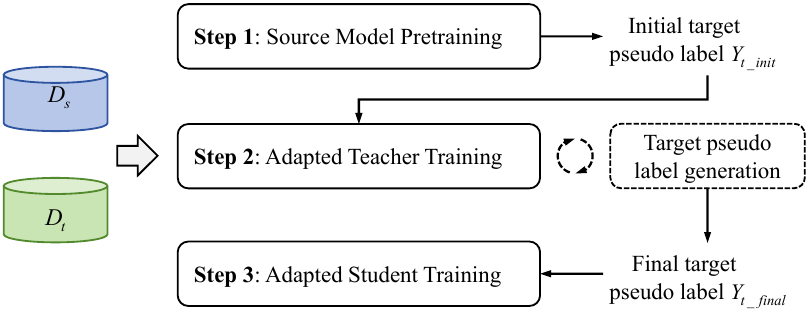}
    \caption{
    % \textcolor{red}{
    Paradigm of self-training in domain adaptive object detection}
    % }
    \label{fig:paradiam}
\end{figure}

\subsection{\textbf{De-simplifying pseudo labels}}
\label{section:my method}

We introduce DeSimPL (De-simplifying pseudo labels) to enhance the model's performance on the target domain by reducing the proportion of simple samples. 
Unsupervised Domain Adaptation (UDA) for object detection often leverages self-training or pseudo-labeling strategies to adapt a model trained on a labeled source domain ($D_s$) to an unlabeled target domain ($D_t$). 
The general paradigm of such approaches, illustrated in Figure~\ref{fig:paradiam}, typically involves several key stages.
Step 1: Source Teacher Pretraining. A teacher model is initially trained on the labeled source domain $D_s$. This model then generates initial pseudo labels ($Y_{t\_init}$) for the target domain $D_t$.
Step 2: Adapted Teacher Training. The teacher model is further trained, often iteratively, using a combination of source data and the 
pseudo-labeled target data. 
During this stage, the target pseudo labels can be progressively refined or updated based on the evolving predictions of the teacher model. 
This iterative process of training and pseudo-label generation is crucial for improving label quality and model adaptation.
Step 3: Adapted Student Training. In many frameworks, particularly those employing a teacher-student architecture, a separate student model is then trained using the refined pseudo labels ($Y_{t\_final}$) generated by the adapted teacher, often in conjunction with the source data. 
In this paper, we take SimROD as our baseline to describe our approach. 
SimROD adopts a teacher-student framework following the above-mentioned diagram. First, the teacher and student models are pre-trained on $D_{s}$. 
Then, the teacher model is fine-tuned on both $D_{s}$ and $D_{t}$ and is used to generate initial pseudo labels on $D_{t}$. Finally, the student model is trained on both $D_{s}$ and $D_{t}$ using the generated pseudo labels to obtain an adapted student model. 
The SimROD with online update approach iteratively updates the pseudo labels in the teacher adaption step to achieve optimal performance.
\par
Our work focuses on enhancing the teacher model in Step 2. As depicted in Figure~\ref{fig:overall_method}, the DeSimPL module consists of three key components: the instance-level memory bank, the data augmentation with adversarial samples, and the adaptive weighted loss. 
In the subsequent sections, we will explicate each module in detail.

\subsubsection{\textbf{Instance-level memory bank}}
\label{section:online_update_strategy}
\begin{figure}[t]
   \begin{center}
   \includegraphics[width=.95\linewidth]{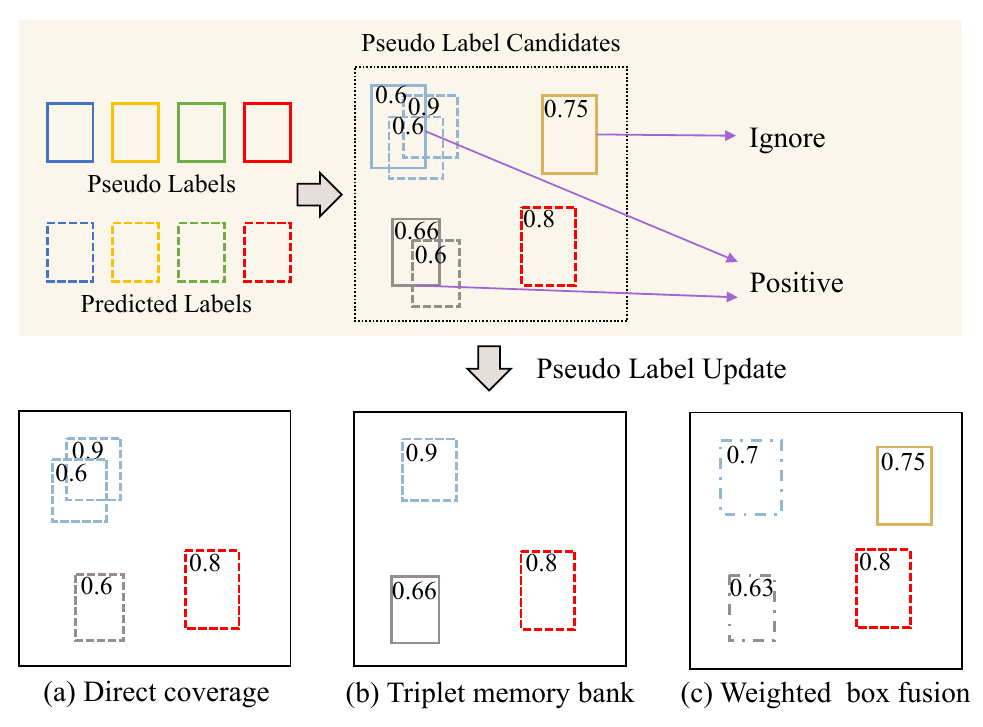}
   \end{center}
   \caption{Comparison of three different ways to update pseudo labels. 
   The number above the box indicates the confidence of the box. 
   Positive and ignore are the labels of the box in the triplet memory bank method (MEV-C) \cite{yang2021st3d}.}
   \label{fig:wbf}
\end{figure}

In recent works, methods for updating pseudo labels can be classified into two categories: direct coverage and pseudo-label fusion. 
The former, exemplified by SoftTeacher \cite{xu2021end}, directly replaces pseudo labels with the latest predictions on the target domain, as shown in Figure~\ref{fig:wbf} (a). 
However, this approach has a drawback: as the model training fluctuates, the performance of the pseudo labels also declines, which further affects subsequent model training. 
The latter category, represented by ST3d \cite{yang2021st3d}, utilizes the triplet memory bank, which combines memory ensemble operation and memory voting to update the pseudo labels, as shown in Figure~\ref{fig:wbf} (b). 
Specifically, the pseudo labels are classified into three types based on their confidence levels: positive, ignore, and discard. 
The positive label serves as a supervisory signal, the ignore label is temporarily reserved and the discard label is directly removed.
When updating pseudo labels, it is necessary to calculate the intersection over union (IoU) between the new and old pseudo labels. 
For a pair of boxes with an IoU greater than a threshold, the box with the higher confidence score should be retained, otherwise it will have a demotion, from positive to ignore, or ignore to discard, if the IoUs between that old box and other boxes are less than the threshold. 
New boxes whose IoUs with the old boxes are lower than the threshold are retained. 
However, this approach is relatively intricate and introduces multiple parameters. Furthermore, retaining only one box from matching pairs of boxes may hinder the fine-tuning of pseudo labels, ultimately resulting in the overfitting of the model to these unchanged pseudo labels.

Following our observations, we have adopted a strategy for preserving the pseudo labels for the target domain images in an instance-level memory bank. 
To effectively integrate the pseudo labels with the model predictions, we employ Weighted Box Fusion (WBF)~\cite{solovyev2021weighted}, as shown in Figure~\ref{fig:wbf} (c). 
\par
Unlike NMS \cite{neubeck2006efficient} and SoftNMS \cite{bodla2017soft}, which directly discard certain predicted boxes, WBF utilizes information from all boxes for fusion, resulting in a more comprehensive integration of pseudo labels. WBF allows for dynamic updates of the pseudo labels while retaining valuable information, ensuring a more precise and robust label refinement process.
The update method is as follows: 
\begin{itemize} 
    \item[1.] To start, we use a pre-trained teacher model from the source domain to generate initial pseudo labels. These labels are filtered with a confidence threshold higher than 0.6 to ensure high precision in the initial pseudo-label set, which is applied to the target domain. \item[2.] During the training process, we apply a confidence threshold of 0.05 to filter the predictions made by the current model on the target domain. We then fuse the filtered results with the pseudo labels from the instance-level memory bank using WBF. To accomplish this, we group all boxes based on IoU and weight and average the coordinates and confidence of the boxes in each cluster, as shown in Figure~\ref{fig:wbf} (c). The fused results are used to update the instance-level memory bank dynamically, ensuring that pseudo labels remain accurate and adaptive throughout training. This process allows the pseudo labels to be continuously fine-tuned while the memory bank's boxes are repeatedly fused with the model's predictions. 
    \item[3.] Finally, we alternate between training the model and updating the pseudo labels using the teacher model. This iterative refinement ensures optimal performance at the end of the training process. This step also reduces the risk of overfitting by maintaining a balance between model predictions and pseudo labels. \end{itemize}

Figure~\ref{fig:wbf} showcases the WBF-based update procedure. Our update strategy prioritizes both precision and recall of pseudo labels. When generating initial pseudo labels, we prioritize high precision since accurate pseudo labels provide a reliable foundation for the teacher model to learn target domain knowledge early in training. Additionally, precise cluster centers can be used for subsequent fusion. For the filtered predictions of the current model on $D_t$, we focus on achieving high recall to complement the initial pseudo labels during the fusion process. This balanced approach ensures that pseudo labels remain representative of the target domain while reducing noise. Furthermore, pseudo label coordinates are continually fine-tuned with WBF to prevent the model from overfitting to noisy pseudo labels.

\subsubsection{\textbf{Data augmentation with adversarial samples}}
\label{section:adversarial_samples}
To achieve superior domain adaptation performance and to mitigate the dominance of simple samples, we integrate two powerful techniques: DomainMix data augmentation \cite{ramamonjison2021simrod} and adversarial examples.
Specifically, the use of DomainMix helps to diversify the training data and increase the model's exposure to various domain-specific features, while the inclusion of adversarial samples encourages the model to learn more robust and discriminative representations by introducing perturbations to the input data and increase the proportion of hard samples in pseudo labels. 
By using the Fast Gradient Sign Attack (FGSM)~\cite{goodfellow2014explaining}, a gradient-based adversarial attack, we can create adversarial samples efficiently and effectively, leading to improved performance and generalization of the model. 
We perform FGSM on the image after DomainMix in accordance with Equation~\ref{equ:fgsm}.
\begin{equation}
   \label{equ:fgsm}
   x' = x + \varepsilon \cdot \mathrm{sign} \{\nabla_xL(x,y)\}
\end{equation}

In the training process, the model is updated using clean images from each batch, followed by a gradient update using adversarial examples from the same batch. Adversarial samples help to address the problem of simple samples, increase the model's ability to adapt to changes in the domain, and improve its performance in the target domain.
\subsubsection{\textbf{Adaptive weighted loss}}
\label{section:adaptive_weighted_loss}
\begin{figure}
   \begin{center}
   \includegraphics[width=.95\linewidth]{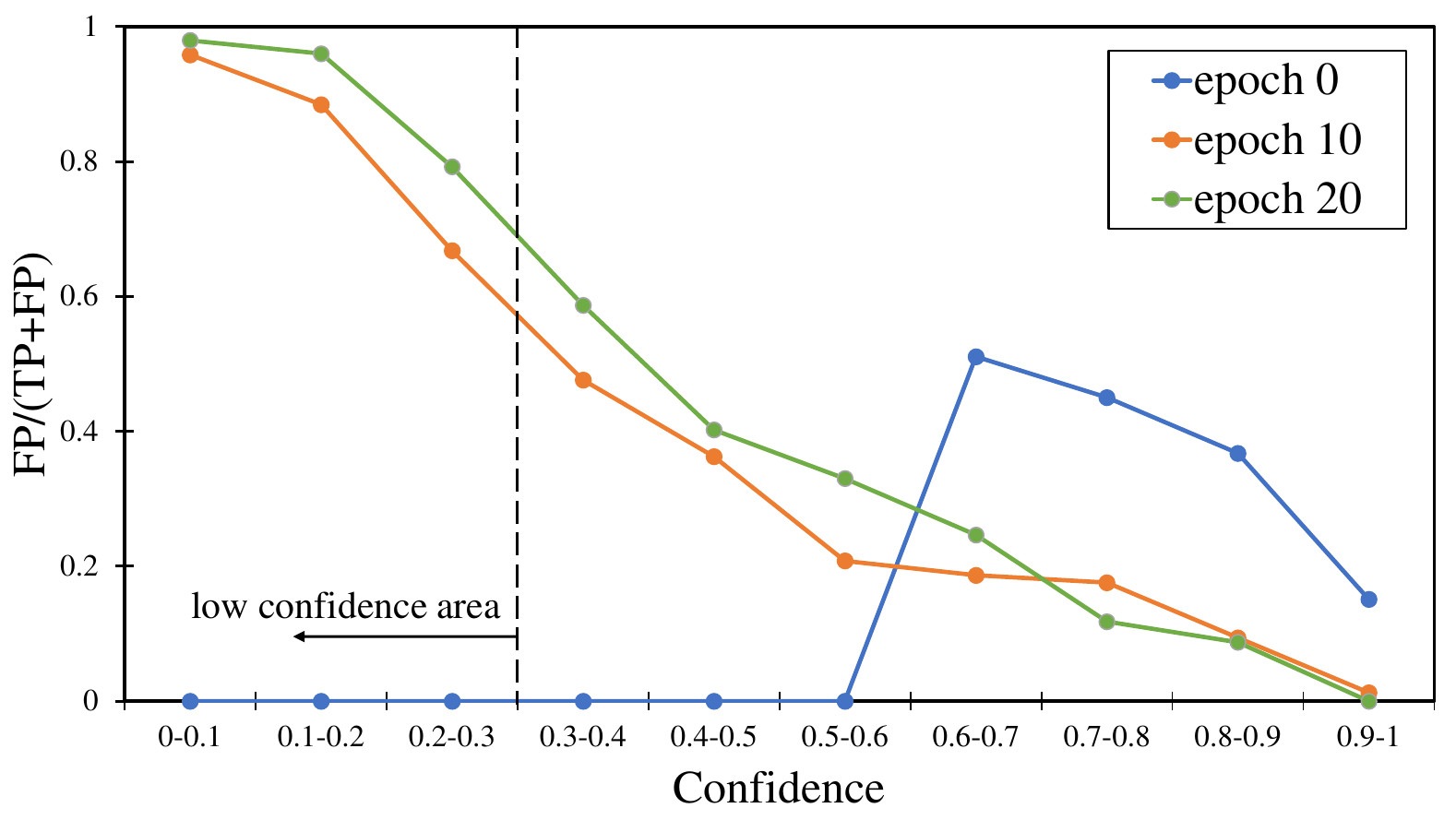}
   \end{center}
      \caption{The relationship between the rate of false positive (FP) pseudo labels and confidence level at different stages of training when using the updated pseudo labels.
      Prior to training, only pseudo labels with a confidence score higher than 0.6 were retained. As training progresses, the proportion of FP in the pseudo labels increases in the low confidence score region, while the proportion of FP in the high confidence score region remains relatively low.}
   \label{fig:scoredistri}
   % \vspace{-1cm}
\end{figure}

During training, the Adaptive Weighted Loss (AWL) is used to mitigate the impact of false positive (FP) pseudo labels by dynamically weighting the localization loss based on pseudo-label confidence. As shown in Figure~\ref{fig:scoredistri}, many FP pseudo labels accumulate in the low-confidence region (confidence $\leq 0.3$) during the later stages of training. By assigning lower weights to low-confidence pseudo labels, the model reduces their influence and focuses on high-confidence labels, improving robustness.
The total loss is formulated as:

\begin{equation}
\mathcal{L}_{\text {total }}=\mathcal{L}_S+\mathcal{L}_T^{\mathrm{cls}}+w \mathcal{L}_T^{\mathrm{loc}}
\end{equation}

Here, $\mathcal{L}_S$ represents the pretraining loss on labeled source domain data, $\mathcal{L}_T^{\text{cls}}$ is the classification loss on target pseudo labels, and $\mathcal{L}_T^{\text{loc}}$ is the localization loss on target pseudo labels. 
Motivated by our observation in Figure~\ref{fig:scoredistri}, we define a confidence threshold $\tau$ to differentiate the weighting. For our experiments, this threshold $\tau$ is set to $0.3$ (the specific implementation of which is detailed in Section IV.A). The weight $w$ is then calculated as follows:
\begin{equation}
w= \begin{cases}1 & \text { if } c>\tau \\ c & \text { if } c \leq \tau\end{cases}
\end{equation}
This dynamic weighting scheme assigns full weight ($w=1$) to high-confidence pseudo labels ($c>\tau$) while retaining the original confidence score as the weight for low-confidence ones ($c \leq \tau$). 
By assigning lower weights to low-confidence pseudo labels for the localization task, the model reduces their influence and focuses on learning from high-confidence, reliable labels, thus improving robustness against noise. This ensures the model effectively balances learning from reliable labels while mitigating noise from less certain ones, enhancing its overall performance in domain adaptation tasks.

\section{Experiments}

\subsection{\textbf{Implementation details}}
Our overall training framework is divided into three parts. For clarity, we explain the implementation details step by step.
\par
\textbf{Step 1: Source-domain pre-training.}
We follow the Source-domain pre-training of SimROD with the single-stage detection model YOLOv5~\cite{glenn_jocher_2020_4154370}, 
Teacher and student are set with YOLOv5x and YOLOv5s, respectively.
The source models are obtained through transfer learning from COCO~\cite{lin2014microsoft} pre-trained weights, following SimROD.
For Pascal, we use a learning rate of $4e^{-5}$ and a batch size of 128. 
For Sim10k and KITTI, we use a learning rate of $4e^{-5}$ and a batch size of 64.
We did not use multi-scale training to simplify our analysis.
Under the adaptation settings Pascal VOC $\rightarrow$ Comic and Pascal VOC $\rightarrow$ Clipart, we resize the training image and test image to 416 pixels.
Under the adaptation settings Sim10k $\rightarrow$ Cityscapes or KTIII $\rightarrow$ Cityscapes, we resize the training and test image to 512 pixels.
\par
\textbf{Step 2: Adapt the teacher model.}
Our proposed method is applied to the adaptation of the teacher model. 
The initial pseudo labels are generated with a confidence threshold of 0.6. 
For WBF in the proposed instance-level memory bank, an IoU threshold euqals 0.5 is used for matching boxes. 
Adversarial samples are generated using an epsilon value of 0.01 as defined in Equation~\ref{equ:fgsm}. 
The adaptive weighted loss function is employed with a confidence threshold of 0.3, setting the confidence of pseudo labels higher than this threshold to 1 while maintaining the confidence of remaining pseudo labels unchanged. 
The experimental setups vary depending on the domain adaptation settings. The learning rate, batch size, number of epochs, and pseudo label update interval are adjusted accordingly. 
For example, when the adaptation setting is Pascal $\rightarrow$ to Comic or Pascal $\rightarrow$ to Clipart, the learning rate is set to $3e^{-5}$, the batch size is 48, and the model is trained for 20 or 10 epochs with pseudo-label updates every 10 or 1 epochs. 
When the adaptation setting is Sim10k $\rightarrow$ Cityscapes or KITTI $\rightarrow$ Cityscapes, the learning rate is set to $1e^{-5}$, the batch size is 16, and the model is trained for 10 or 20 epochs with pseudo-label updates every 1 or 10 epochs, respectively. 
It is worth noting that when the pseudo-label update interval is greater than 1 epoch, the pseudo labels are updated once after the first epoch of model training and then updated according to the update interval.
\par
\textbf{Step 3: Adapt the student model.}
In this step, we train the student model following SimROD.
We first use the trained teacher model to generate pseudo labels and then adapt the student model. 
When the student model is adapted from Pascal VOC2007 to the Comic, the confidence threshold is set to 0.2, the learning rate is set to $6e^{-5}$, and the batch size is set to 96. 
For other domain adaptation settings, the confidence threshold is set to 0.3, the learning rate is set to $4e^{-5}$, and the batch size is set to 64.
In addition, the total epochs are 200.

\subsection{\textbf{Datasets}}
In the experiments, we use six object detection datasets: Pascal VOC \cite{everingham2009pascal}, Comic \cite{inoue2018cross}, Clipart \cite{inoue2018cross}, Sim10k \cite{johnson2016driving}, Cityscapes \cite{cordts2016cityscapes}, and KITTI \cite{geiger2012we}. 
As shown in Table~\ref{tab:datasets_illustration}, Pascal VOC includes VOC 2007 (including 5k images) and VOC 2012 (including 11k images), a total of 16,551 real-scene images, and 20 categories. 
Comic is a cartoon dataset that contains 1k training images and 1k test images and shares 6 categories with Pascal VOC. 
Clipart and Pascal VOC have the same 20 categories, including 1k images. 
Sim10k is the simulation scene image dataset, which contains 10k training images and 58,701 car category labeling information. 
Cityscapes have 2,975 training images and 500 validation images, with a total of 8 categories. 
KITTI contains 7,481 images, following prior works~\cite{li2022sigma}, only the car class is used.

\begin{table}[!t]
 \caption{Summary of datasets used in our domain adaptive object detection experiments.}
\centering
\begin{tabular}{lcccc}
\Xhline{1.4pt}
\multicolumn{1}{c}{\multirow{2}{*}{Dataset}} & \multicolumn{2}{c}{Train} & \multicolumn{2}{c}{Test} \\
\cline{2-3}
\cline{4-5}
\multicolumn{1}{c}{}                         & Images    & Catagories    & Images    & Catagories   \\
\hline
Pascal VOC 2007                              & 5011      & 20            & 5011      & 20           \\
Pascal VOC 2012                              & 11540     & 20            & 11540     & 20           \\
Comic                                        & 1000      & 6             & 1000      & 6            \\
Clipart                                      & 1000      & 20            & 1000      & 20           \\
Sim10k                                       & 10000     & 1             & 10000     & 1            \\
Cityscapes                                   & 2975      & 8             & 500       & 8            \\
KITTI                                        & 7481      & 1             & 7481      & 1
\\
\Xhline{1.4pt}         
\end{tabular}
\label{tab:datasets_illustration}
\end{table}

\begin{table*}[!t]
\vspace{0.3cm}
 \caption{Results on Real (VOC) $\rightarrow$ Clipart. ``R101'' represent the ResNet101 backbones.
   ``S416'',``X416'' represents different scales of YoloV5 model. ``Source'' represents the performance of the model trained only on source images.
   We report the mAP50 (\%) performance of the adapted model.}
   \begin{center}
   
   \resizebox{\linewidth}{!}{
      \begin{tabular}{l|l|c|cccccccccccccccccccc|c|c}
      % \toprule
      \Xhline{1.4pt}
      Method & Arch. & Backbone & aero & bcycle & bird & boat & bottle & bus & car & cat & chair & cow & table & dog & hrs & bike & prsn & plnt & sheep & sofa & train & tv & mAP50 & Source \\
      \hline

      DAF \cite{chen2018domain} & F-RCNN & R101 & 15.0 & 34.6 & 12.4 & 11.9 & 19.8 & 21.1 & 23.2 & 3.1 & 22.1 & 26.3 & 10.6 & 10.0 & 19.6 & 39.4 & 34.6 & 29.3 &1.0 & 17.1 & 19.7 & 24.8 & 19.8 & 27.8\\
      BDC-Faster \cite{saito2019strong} & F-RCNN & R101 & 20.2 & 46.4 & 10.4 & 19.3 & 18.7 & 41.3 & 26.5 & 6.4 & 33.2 & 11.7 & 26.0 & 1.7 & 36.6 & 41.5 & 37.7 & 44.5 & 10.6 & 20.4 & 33.3 & 15.5 & 25.6 & 27.8\\
      WST-BSR \cite{kim2019self} & F-RCNN & R101 & 28.0 & 64.5 & 23.9 & 19.0 & 21.9 & 64.3 & 43.5 & 16.4 & 42.0 & 25.9 & 30.5 & 7.9 & 25.5 & 67.6 & 54.5& 36.4 & 10.3 & 31.2 & 57.4 & 43.5 & 35.7 & 27.8\\
      SWDA \cite{saito2019strong} & F-RCNN & R101 & 26.2 & 48.5 & 32.6 & 33.7 & 38.5 & 54.3 & 37.1 & 18.6 & 34.8 & 58.3 & 17.0 & 12.5 & 33.8 & 65.5 & 61.6 & 52.0 & 9.3 & 24.9 &54.1 & 49.1 & 38.1 & 27.8\\
      MAF \cite{he2019multi} &  F-RCNN & R101 & 38.1 & 61.1 & 25.8 & 43.9 & 40.3 & 41.6 & 40.3 & 9.2 & 37.1 & 48.4 & 24.2 & 13.4 & 36.4 & 52.7 & 57.0 & 52.5 & 18.2 & 24.3 & 32.9 & 39.3 & 36.8 & 27.8\\
      SCL \cite{shen2019scl} &  F-RCNN & R101 & 44.7 & 50.0 & 33.6 & 27.4 & 42.2 & 55.6 & 38.3 & 19.2 & 37.9 & \textbf{69.0} & 30.1 & \textbf{26.3} & 34.4 & 67.3 & 61.0 & 47.9 & 21.4 & 26.3 & 50.1 & 47.3 & 41.5 & 27.8\\
      CRDA \cite{xu2020exploring} & F-RCNN & R101 & 28.7 & 55.3 & 31.8 & 26.0 & 40.1 & 63.6 & 36.6 & 9.4 & 38.7 & 49.3 & 17.6 & 14.1 & 33.3 & 74.3 & 61.3 & 46.3 & 22.3 & 24.3 & 49.1 & 44.3 & 38.3 & 27.8\\
      HTCN \cite{chen2020harmonizing} & F-RCNN & R101 & 33.6 & 58.9 & 34.0 & 23.4 & 45.6 & 57.0 & 39.8 & 12.0 & 39.7 & 51.3 & 21.1 & 20.1 & 39.1 & 72.8 & 63.0 & 43.1 & 19.3 & 30.1 & 50.2 & 51.8 & 40.3 & 27.8\\
      ATF \cite{he2020domain} &  F-RCNN & R101 & 41.9 & 67.0 & 27.4 & 36.4 & 41.0 & 48.5 & 42.0 & 13.1 & 39.2 & 75.1 & \textbf{33.4} & 7.9 & 41.2 & 56.2 & 61.4 & 50.6 & \textbf{42.0} & 25.0 & 53.1 & 39.1 & 42.1 & 27.8\\
      Unbiased \cite{liu2021unbiased} &F-RCNN&R101& 30.9 & 51.8 & 27.2 & 28.0 & 31.4 & 59.0 & 34.2 & 10.0 & 35.1 & 19.6 & 15.8 & 9.3 & 41.6 & 54.4 & 52.6 & 40.3 & 22.7 & 28.8 & 37.8 & 41.4 & 33.6 & 27.8\\

      D-adapt~\cite{jiangdecoupled} & F-RCNN&R101& \textbf{56.4} & 63.2 & \textbf{42.3} & 40.9 & 45.3 & 77.0 & 48.7 & 25.4 & 44.3 & 58.4 & 31.4 & 24.5 & \textbf{47.1} & 75.3 & 69.3 & 43.5 & 27.9 & 34.1 & 60.7 & \textbf{64.0} & \textbf{49.0} & 27.8\\
      \hline
      % \midrule
      SIGMA~\cite{li2022sigma} & FCOS & R101 & 40.1 & 55.4& 37.4 & 31.1 & 54.9 & 54.3 & 46.6 & 23.0 & 44.7 & 65.6 & 23.0 & 22.0 & 42.8 & 55.6 & 67.2 & 55.2& 32.9 & 40.8 & 45.0 & 58.6 & 44.5 & 25.3 \\
      SIGMA++~\cite{li2023sigma++} & FCOS & R101 & 36.3 & 54.6 & 40.1 & 31.6 & \textbf{58.0} & 60.4 & 46.2 & \textbf{33.6} & 44.4 & 66.2 & 25.7 & 25.3 & 44.4 & 58.8 & 64.8 & 55.4 & 36.2 & 38.6 & 54.1 & 59.3 & \textbf{46.7} & 25.3 \\
      \hline
      % \midrule
      \begin{tabular}[l]{@{}l@{}}SimROD \\ \scriptsize{\color{black}{(w. teacher X416)}} \end{tabular} & YOLOv5 & S416 & 40.5 & 74.1 & 40.0 & 41.4 & 53.8 & 81.9 & 64.7 & 7.8 & 66.7 & 50.9 & 17.7 & 10.0 & 42.8 & 60.3 & 76.4 & 63.1 & 19.4 & 42.7 & 64.1 &60.1& 48.9 & 27.1\\
      \begin{tabular}[l]{@{}l@{}}Ours \\\scriptsize{\color{black}{(w. teacher X416)} }\end{tabular} & YOLOv5 & S416 & 33.1 & \textbf{77.4} & \textbf{50.6} & \textbf{47.8} & 56.1 & \textbf{87.3} & \textbf{71.2} & 11.6 & \textbf{67.6} & 60.0 & 26.2 & 12.8 & 44.9 & \textbf{80.4} & \textbf{84.0} & \textbf{62.2} & 24.6 & \textbf{47.0} & \textbf{67.1} & 59.6 & \textbf{53.6} & 27.1\\
      % \bottomrule
      \Xhline{1.4pt}
      \end{tabular}
   }
   \end{center}
   \label{tab:voc0712_clipart}
\end{table*}
\begin{table*}[!t]
\caption{Results on Real (VOC) $\rightarrow$ Comic. ``V'' represents the VGG16 backbone.
   ``S416'', ``X416'' represents different scales of YoloV5 model. ``Source'' represents the performance of the model trained only on source images.
   ``Oracle'' represents the performance of the model trained on labeled target data.
   We report the mAP50 (\%) performance of the adapted model.}
   \begin{center}
   % \resizebox{1.00\columnwidth}{!}{
   \setlength{\tabcolsep}{12.7pt}{
      \begin{tabular}{l|l|c|c|c|c}
      % \hline
      \Xhline{1.4pt}
      Method & Arch. & Backbone & Source & mAP50 & Oracle \\
      \hline
      % \hline
      ADDA \cite{tzeng2017adversarial} & SSD & V & 24.9 & 23.8 & 46.4 \\
      DT \cite{inoue2018cross} & SSD & V & 24.9 & 29.8 & 46.4 \\
      DT+PL \cite{inoue2018cross} & SSD & V & 24.9 & \textbf{37.2} & 46.4 \\
      \hline
      DAF \cite{chen2018domain} & F-RCNN & V & 21.4 & 23.2 & - \\
      DT \cite{inoue2018cross} & F-RCNN & V & 21.4 & 29.8 & - \\
      SWDA \cite{saito2019strong} & F-RCNN & V & 21.4 & 28.4 & - \\
      DAM \cite{kim2019diversify} & F-RCNN & V & 21.4 & \textbf{34.5} & - \\
      \hline
      DeepAugment \cite{hendrycks2021many} & YOLOv5 & S416 & 18.2 & 21.4 & 39.8\\
      BN-Adapt \cite{ioffe2015batch} & YOLOv5 & S416 & 18.2 & 25.5 & 39.8\\
      Stylize \cite{geirhos2018imagenet} & YOLOv5 & S416 & 18.2 & 27.6 & 39.8 \\
      STAC \cite{sohn2020simple} & YOLOv5 & S416 & 18.2 & 26.4 & 39.8\\
      DT+PL \cite{inoue2018cross} & YOLOv5 & S416 & 18.2 & 25.7 & 39.8\\
      SimROD (w. teacher X416) 

      & YOLOv5 & S416 & 18.2 & 37.6 & 39.8\\
      Ours (w. teacher X416) & YOLOv5 & S416 & 18.2 & \textbf{39.5} & 39.8\\

      % \hline
      \Xhline{1.4pt}
      \end{tabular}
   }
   \end{center}
   
   \label{tab:voc2comic}
\end{table*}
\begin{table*}[!t]
   \caption{Results on Sim10k $\rightarrow$ Cityscapes. ``V'', ``I'', ``R50'' and ``R101'' represent the VGG16, Inception-v2, ResNet50 and ResNet101 backbones respectively.
   ``S512'', ``S416'', ``X512'' and ``X1280'' represents different scales of YoloV5 model. ``Source'' represents the performance of the model trained only on source images.
   ``Oracle'' represents the performance of the model trained on labeled target data.
   ``*'' denotes this method utilizes CycleGAN \cite{zhu2017unpaired} to perform source-to-target translation.
   We report the AP50 (\%) performance of the adapted model.}
   \begin{center}
   % \resizebox{1.00\columnwidth}{!}{
   \setlength{\tabcolsep}{12.7pt}{
      \begin{tabular}{l|l|c|c|c|c}
      % \hline
      \Xhline{1.4pt}
      Method & Arch. & Backbone & Source & AP on Car & Oracle \\
      \hline
      % \hline
      DAF \cite{chen2018domain} & F-RCNN & V & 30.1 & 39.0 & - \\
      MAF \cite{he2019multi} & F-RCNN & V & 30.1 & 41.1 & - \\
      RLDA \cite{khodabandeh2019robust} & F-RCNN & I & 31.1 & \textbf{42.6} & 68.1 \\
      \hline
      SCDA \cite{zhu2019adapting} & F-RCNN & V & 34.0 & 43.0 & - \\
      MDA \cite{xie2019multi} & F-RCNN & V & 34.3 & 42.8 & - \\
      SWDA \cite{saito2019strong} & F-RCNN & V & 34.6 & 42.3 & - \\
      Selective DA \cite{zhu2019adapting} & F-RCNN & V & 34.6 & 43.0 & 69.7\\
      CDN \cite{su2020adapting} & F-RCNN & V & 34.6 & 43.9 & 69.7 \\
      HTCN* \cite{chen2020harmonizing} & F-RCNN & V & 34.6 & 42.5 & 69.7 \\
      ATF \cite{he2020domain} & F-RCNN & V & 34.6 & 42.8 & 69.7 \\
      MeGA-CDA \cite{vs2021mega} & F-RCNN & V & 34.6 & \textbf{44.8} & 69.7 \\
      UMT* \cite{deng2021unbiased} & F-RCNN & V & 34.6 & 43.1 & 69.7 \\
      Coarse-to-Fine \cite{zheng2020cross} & F-RCNN & V & 35.0 & 43.8 & 59.9 \\
      
      \hline
      MTOR \cite{cai2019exploring} & F-RCNN & R50 & 39.4 & 46.6 & - \\
      
      ViSGA  \cite{rezaeianaran2021seeking} & F-RCNN & R50 & 39.4 & 49.3 & -\\ 

      D-adapt \cite{jiangdecoupled} & F-RCNN & R101 & 41.8 & 51.9 & 70.4 \\
      EveryPixelMatters \cite{hsu2020every} & FCOS & V & 39.8 & 49.0 & 69.7 \\
      SIGMA \cite{li2022sigma} & FCOS & V & 39.8 & 53.7 & - \\ 

      SIGMA$++$  \cite{li2023sigma++} & FCOS & V & 39.8 & \textbf{57.7} & - \\
      SimROD(w. teacher X512) 
      & YOLOv5 & S512 & 44.4 & 53.5 & 58.3 \\
      Ours (w. teacher X512) & YOLOv5 & S512 & 44.4 & \underline{55.3} & 58.3 \\

      % \hline
      \Xhline{1.4pt}
      \end{tabular}
   }
   \end{center}

   \label{tab:sim10k2city}
\end{table*}
% \clearpage
% \vspace{1cm}
\begin{table*}[!t]
\caption{Results on KITTI $\rightarrow$ Cityscapes. ``V'', ``I'' and ``R50'' represent the VGG16, Inception-v2 and ResNet50 backbones respectively.
   ``S512'', ``S416'', ``X512'' and ``X1280'' represents different scales of YoloV5 model. ``Source'' represents the performance of the model trained only on source images.
   ``Oracle'' represents the performance of the model trained on labeled target data.
   We report the AP50 (\%) performance of the adapted model.}
   \begin{center}
   % \resizebox{1.00\columnwidth}{!}{
   \setlength{\tabcolsep}{12.7pt}{
      \begin{tabular}{l|l|c|c|c|c}
      % \hline
      \Xhline{1.4pt}
      Method & Arch. & Backbone & Source & AP on Car & Oracle \\
      \hline
      % \hline
      DAF \cite{chen2018domain} & F-RCNN & V & 30.2 & 38.5 & - \\
      MAF \cite{he2019multi} & F-RCNN & V & 30.2 & 41.0 & - \\
      RLDA \cite{khodabandeh2019robust} & F-RCNN & I & 31.1 & 43.0 & 68.1 \\
      MeGA-CDA \cite{vs2021mega} & F-RCNN & V & 30.2 & \textbf{43.0} & - \\
      \hline
      
      SCDA \cite{zhu2019adapting} & F-RCNN & V & 37.4 & 42.6 & - \\
      ViSGA  \cite{rezaeianaran2021seeking} & F-RCNN & R50 & 32.5 & 47.6 & - \\
      EveryPixelMatters \cite{hsu2020every} & FCOS & R50 & 35.3 & 45.0 & 70.4 \\
      KTNet \cite{tian2021knowledge} & FCOS & V & 34.4 & 45.6 & - \\
      SSAL \cite{munir2021ssal} & FCOS & V & 34.9 & 45.6 & - \\
      SIGMA \cite{li2022sigma} & FCOS & V & 34.4 & 45.8 & - \\
      SIGMA$++$  \cite{li2023sigma++} & FCOS & V & 34.4 & 49.5 & - \\
      SimROD (w. teacher X512)  
      & YOLOv5 & S512 & 38.5 & 50.3 & 58.5 \\
      Ours(w. teacher X512) & YOLOv5 & S512 & 38.5 & \textbf{52.1} & 58.5 \\   
      % \hline
      \Xhline{1.4pt}
      \end{tabular}
   }
   \end{center}
   
   \label{tab:kitti2city}
\end{table*}
\begin{table}[t!]
 \caption{Experiment about adaptive weighted loss on VOC to Comic with YOLOv5x.  ``wbox'' represents adaptive weighting of location loss.
   ``wcls'' represents the adaptive weighting of classification loss. We report the mAP50 (\%) performance of the adapted model.}
   \begin{center}
   % \resizebox{1.00\columnwidth}{!}{
   \setlength{\tabcolsep}{12.7pt}{
      \begin{tabular}{l|l|c|c}
      % \hline
      \Xhline{1.4pt}
      Method & Arch. & Backbone & mAP50 \\
      \hline
      % \hline
      no weighted loss & YOLOv5 & X416 & 48.6 \\ 
      wbox & YOLOv5 & X416 & 49.0 \\
      wbox \& wcls & YOLOv5 & X416 & 48.6 \\
      % \hline
      \Xhline{1.4pt}
      \end{tabular}
   }
   \end{center}
   
   \label{tab:weighted_loss}
\end{table}
\begin{table}[t!]
\caption{The impact of $\varepsilon$ in Eq.1 on model performance under the setting Real (VOC) $\rightarrow$ Clipart.}
\centering
\setlength{\tabcolsep}{12.7pt}{
\begin{tabular}{l|cccc}
\Xhline{1.4pt}
$\varepsilon$ & Arch.  & Backbone & mAP50 &  \\ \hline
0          & YOLOv5 & X416     & 49.0  &  \\
0.01       & YOLOv5 & X416     & 51.1  &  \\
0.05       & YOLOv5 & X416     & 47.8  &  \\
0.1        & YOLOv5 & X416     & 45.5  &  \\ \Xhline{1.4pt}
\end{tabular}
}
\label{tab:eps}
\end{table}

\subsection{\textbf{Comparison with state-of-the-arts}}
In this section, we conducted experiments on the two prevalent domain adaptation settings in traffic and transportation scenarios: synthetic to real setting and cross-camera setting. Additionally, we introduced a dissimilar domains setting to offer further validation of the effectiveness of our method.
For the adaptation experiments on Pascal VOC $\rightarrow$ Comic, we use the same data partitioning as SimROD. For the rest of the experiments, the data partitioning is consistent with the mainstream works~\cite{chen2018domain,jiangdecoupled}.
\par
\textbf{Synthetic to real.}
We conducted experiments on domain adaptation from synthetic to real, namely Sim10k $\rightarrow$ Cityscapes. As shown in Table~\ref{tab:sim10k2city}, our method outperforms most other methods and is closest to the Oracle performance compared to D-adapt. This indicates the high accuracy of the pseudo labels generated by our method.
\par
\textbf{Cross camera.}
KITTI dataset is a collection of real-world traffic scenes captured by car-mounted cameras, which results in a domain gap with Cityscapes (on-board cameras). As presented in Table~\ref{tab:kitti2city}, our results exceed all other methods, and we achieve a performance improvement of 2.6\% AP compared to SIGMA++.
\par
\textbf{Dissilimar domains.}
To further validate the effectiveness of our proposed method,we present the adaptation results for dissimilar domains by adapting the model from Pascal VOC2007+2012 to Clipart dataset. Our proposed approach, as shown in Table~\ref{tab:voc0712_clipart}, outperforms the state-of-the-art (SOTA) by 4.6\% mAP. The traditional semi-supervised algorithm Unbiased performs poorly due to inaccurate pseudo labels generated by domain shift. The results demonstrate the effectiveness of our method in generating precise pseudo labels even across different domains.
\par
In addition, we conducted an experiment on Pascal VOC2007 $\rightarrow$ Comic for a detailed comparison with SimROD and ablation study. As shown in Table~\ref{tab:voc2comic}, our proposed approach achieved new SOTA results on AP50, surpassing DT+PL and SimROD by 2.3\% and 1.9\%, respectively, after incorporating DeSimPL in the teacher model's training phase. 

\begin{table}[t]
\caption{
The impact of the confidence threshold during training under the setting Pascal VOC $\rightarrow$ Comic.
}
\centering
\setlength{\tabcolsep}{12.7pt}{
\begin{tabular}{l|ccc}
\Xhline{1.4pt}
Threshold & Arch.  & Backbone & mAP50 \\ \hline
0.01      & YOLOv5 & X416     & 40.2  \\
0.05      & YOLOv5 & X416     & 47.8  \\
0.1       & YOLOv5 & X416     & 47.3  \\
0.3       & YOLOv5 & X416     & 47.3  \\
\Xhline{1.4pt}
\end{tabular}
}
\label{tab:cfd_threshold}
\end{table}

\subsection{\textbf{Ablation study}}
In this section, we present an ablation study with YOLOv5x on Pascal VOC2007 $\rightarrow$ Comic to demonstrate the efficacy of the proposed three components, as shown in Table~\ref{tab:ablation}. Initially, we obtain the pre-trained teacher model from the source domain and then employ it for domain adaptation. Finally, we evaluate the performance of the teacher model under different experimental settings.

\textbf{Ablation on online update pseudo-label strategy.}
Table~\ref{tab:ablation} shows the ablation study with the teacher model (YOLOv5x) on four benchmarks.
On the other hand, the experimental results in the fourth row show that when the proposed instance-level memory bank based online update strategy is combined with SimROD, the performance increases. 
This experimental result demonstrates the effectiveness of our proposed online update strategy.

\textbf{Ablation on adaptive weighted loss.}
The effectiveness of adaptive weighted loss can be observed from the significant improvement in the model's performance as depicted in the fourth and fifth rows of Table~\ref{tab:ablation}. 
Furthermore, Table~\ref{tab:weighted_loss} indicates that the model's performance deteriorates when the calculation of classification loss is weighted according to pseudo-label confidence, while location loss weighting leads to further performance enhancement. 
This experimental finding also demonstrates that the classification accuracy of pseudo-labels is higher compared to their localization accuracy.

\textbf{Ablation on the impact of $\varepsilon$ in Equation~\ref{equ:fgsm}.}
As shown in Table~\ref{tab:eps}, it is evident that when $\varepsilon$ is excessively large, it leads to image contamination by noise, thereby hindering the model's ability to recognize the images.
When $\varepsilon$ is set to 0.01, it generates images with minimal noise, resulting in optimal performance of the model.
Ablations of hyperparameters have been evaluated and will be included in the supplementary material.
In addition, for clarification, the Localization loss of each pseudo label is weighted by its confidence $w$.

\par
\textbf{Ablation on adversarial samples.}
As shown in the last two rows of Table~\ref{tab:ablation}, it is evident that integrating adversarial samples during the training phase can effectively boost the performance of the model, resulting in a notable increase of the mAP from 49.0\% to 51.1\%. 
This outcome highlights the effectiveness of the training strategy that incorporates adversarial samples, which can improve the model's generalization and stability.

\textbf{Ablation on the impact of the confidence threshold during training}
The results in Table~\ref{tab:cfd_threshold} (Pascal VOC $\rightarrow$ Comic) provide valuable insights into the impact of confidence thresholds on pseudo-label filtering. The experiments show that a threshold of 0.05 achieves the best performance, with an mAP of 47.8\%, outperforming both lower and higher thresholds. Specifically, a threshold of 0.01 retains excessive noise, leading to significantly lower performance (40.2\% mAP), while higher thresholds, such as 0.1 or 0.3, exclude too many predictions, resulting in reduced training diversity and lower performance (47.3\% mAP).
These findings demonstrate that overly low thresholds introduce noise into the training process, while overly high thresholds reduce the number of usable pseudo-labels, limiting the model’s ability to adapt to the target domain. The optimal threshold of 0.05 strikes the right balance, effectively filtering noise while retaining sufficient pseudo-label diversity for robust training.

\subsection{\textbf{Analysis}}
In this section, we evaluate the effectiveness of our approach by analyzing the pseudo labels during the teacher model's adaptation process. 
Our experiments were conducted on the Pascal VOC2007 $\rightarrow$ Comic dataset using the YOLOv5x model with an image size of 416 for both training and testing. 
We compare our approach with two others: SimROD~\cite{ramamonjison2021simrod} and SimROD integrated with ST3d’s online update pseudo label method~\cite{yang2021st3d} (i.e., SimROD w. online update). 
SimROD takes 100 epochs to converge, with pseudo labels updated once after 50 epochs of the training. 
The confidence threshold for filtering pseudo labels was set to 0.4 according to SimROD. 
The second approach converges in just 30 epochs, with only positive pseudo labels considered for evaluation. 
Our approach, on the other hand, converges in 20 epochs and uses a confidence threshold of 0.6 for filtering initial pseudo labels. 
We update pseudo labels once after the first epoch of model training, and then every 10 epochs thereafter.
\begin{table*}[t!]
\caption{Ablation study with the teacher model (YOLOv5x). 
   ``ILMB'', ``AWL'' and ``ADV'' are three components of our method. 
   ``MEV-C'' represents the classic pseudo-label update algorithm in \cite{yang2021st3d}. 
   We report the mAP50 (\%) performance of the adapted model.}
   \begin{center} 
      \setlength{\tabcolsep}{8pt}{
      \begin{tabular}{l|ccc|llll} 
      % \hline
      \Xhline{1.4pt}
   Method               & LIMB & AWL & ADV & \begin{tabular}[c]{@{}c@{}}VOC~$\rightarrow$Comic\end{tabular} & \begin{tabular}[c]{@{}c@{}}VOC~$\rightarrow$CliPart\end{tabular} & \begin{tabular}[c]{@{}c@{}}Sim10k~$\rightarrow$Cityscapes\end{tabular} & \begin{tabular}[c]{@{}c@{}}KITTI$~\rightarrow$Cityscapes\end{tabular} \\ 
   \hline
   % \hline
   Source               &      &     &     & 32.8                                                & 32.8                                                   & 56.3                                                        & 51.0                                                       \\
   SimROD             &      &     &     & 46.5                                                & 58.9                                                   & 56.8                                                        & 53.2                                                       \\
   SimROD w. MEV-C    &      &     &     & 46.1\scriptsize{\color{red}{-0.4}}                             & 58.4\scriptsize{\color{red}{-0.5}}                                & 57.5\scriptsize{\color{green}{+0.7}}                                             & 53.3\scriptsize{\color{green}{+0.1}}                                    \\
   \hline
   SimROD  &   \checkmark   &     &     & 48.6\scriptsize{\color{green}{+2.1}}                             & 60.9\scriptsize{\color{green}{+2.0}}                                     & 58.0\scriptsize{\color{green}{+1.2}}                                                    & 53.8\scriptsize{\color{green}{+0.6}}                                     \\
   SimROD &   \checkmark   &  \checkmark   &     & 49.0\scriptsize{\color{green}{+2.5}}                             & 61.1\scriptsize{\color{green}{+2.2}}                                     & 58.3\scriptsize{\color{green}{+1.5}}                                                  & 54.1\scriptsize{\color{green}{+0.9}}                                                   \\ 
   \hline
   \textbf{Ours}        & \checkmark     &  \checkmark   &  \checkmark    & \textbf{51.1\scriptsize{\color{green}{+4.6}}}                                            & \textbf{64.0\scriptsize{\color{green}{+5.1}}}                                               & \textbf{58.7\scriptsize{\color{green}{+1.9}}}                                                  & \textbf{54.4\scriptsize{\color{green}{+1.2}}}                                    \\
   % \hline
   \Xhline{1.4pt}
   \end{tabular}
   }
   % }
   \end{center}
   
   \label{tab:ablation}
\end{table*}

\textbf{Pseudo-label performance analysis.}
For the initial pseudo-label, as we use a higher filtering threshold than SimROD, our method has the worst initial pseudo-label performance, as shown in Figure~\ref{fig:simple_samples_increase}. 
It is worth noting that the pseudo labels in our approach have shown exceptional performance. After just one epoch of training, they achieved a score of 35.3\% mAP.Furthermore, these labels continue to show improvement as the model undergoes further training, eventually reaching an impressive 38.8\% mAP. 
Although SimROD w. online update offers some additional improvement in the performance of pseudo labels, it still falls short compared to the performance achieved using our method. Overall, these results suggest that our approach has the potential for improving the performance of pseudo labels in training deep models.

\begin{table}[t]
\caption{
% \textcolor{red}{
Performance across different cities in cityscapes.
% }
}
\centering
\setlength{\tabcolsep}{12.7pt}{
\begin{tabular}{l|ccc}
\Xhline{1.4pt}
City      & Arch.  & Backbone & mAP50 \\
\hline
All       & YOLOv5 & S512     & 55.3  \\
Frankfurt & YOLOv5 & S512     & 53.3  \\
Lindau    & YOLOv5 & S512     & 67.6  \\
Munster   & YOLOv5 & S512     & 58 \\ \Xhline{1.4pt} 
\end{tabular}
}
\label{tab:different_citie}
\end{table}

\begin{table}[t]
\caption{
% \textcolor{red}{
Performance under foggy conditions.
% }
}
\centering
\setlength{\tabcolsep}{12.7pt}{
\begin{tabular}{l|ccc}
\Xhline{1.4pt}
Foggy\_beta & Arch.  & Backbone & mAP50 \\
\hline
0           & YOLOv5 & S512     & 55.3  \\
0.005       & YOLOv5 & S512     & 52.6  \\
0.01        & YOLOv5 & S512     & 47.9  \\
0.02        & YOLOv5 & S512     & 40.2 \\\Xhline{1.4pt} 
\end{tabular}
}
\label{tab:foggy_conditions}
\end{table}

\begin{figure*}[!t]
   \begin{center}
   \includegraphics[width=0.96\linewidth]{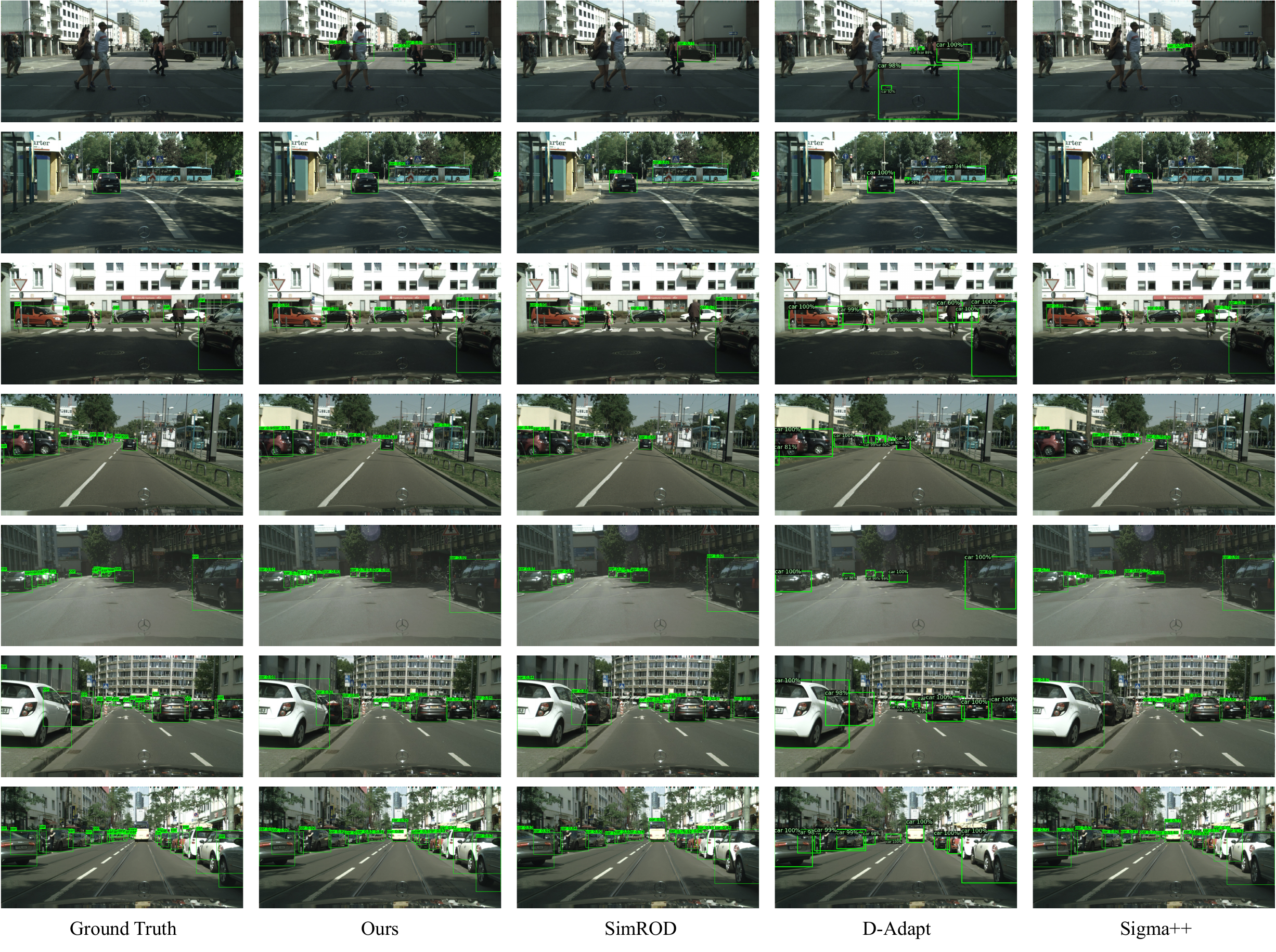}
   \end{center}
   \caption{
   % \textcolor{red}{
   Qualitative comparisons under the setting Sim10k $\rightarrow$ Cityscapes with the YOLOv5 model.}
   % }
   \label{fig:result_vis1}
\end{figure*}

\begin{figure*}[!t]
   \begin{center}
   \includegraphics[width=0.7\linewidth]{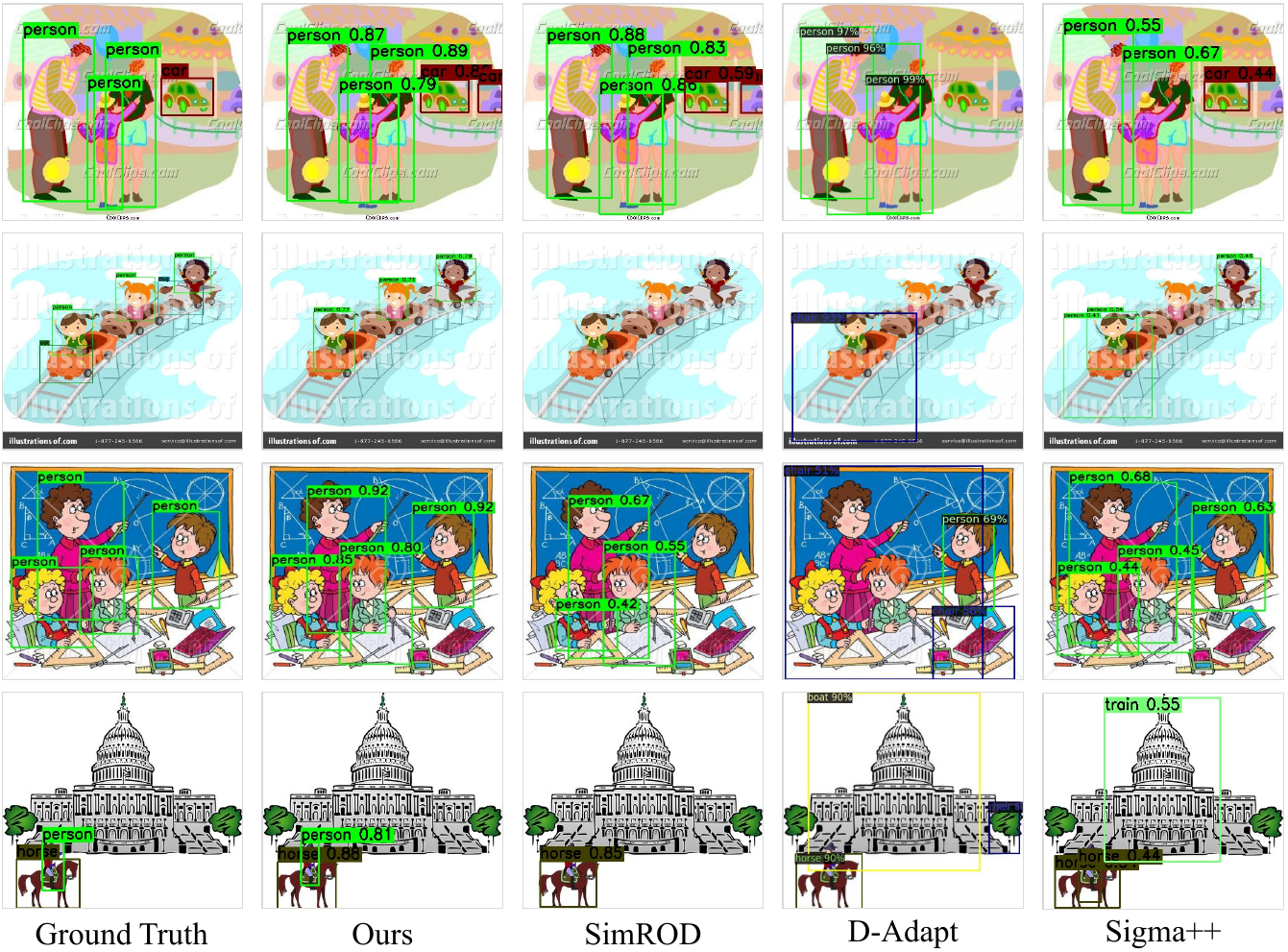}
   \end{center}
   \caption{
   % \textcolor{red}{
   Qualitative comparisons under the setting Pascal VOC $\rightarrow$ Clipart with the YOLOv5 model.
   % }
   }
   \label{fig:result_vis2}
\end{figure*}

\begin{figure*}[!t]
   \begin{center}
   \includegraphics[width=0.96\linewidth]
   {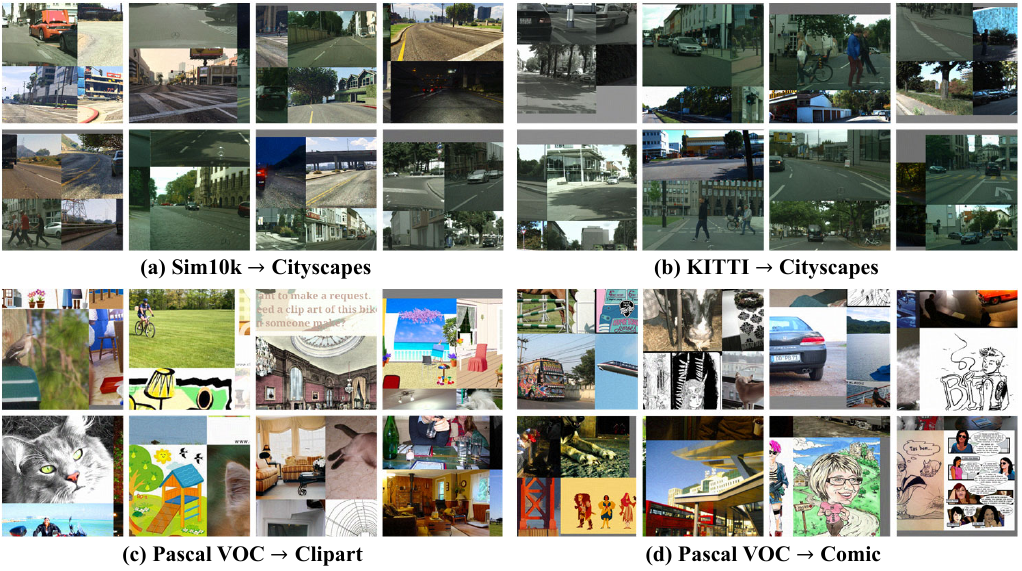}
   % {images/adv_sample_vis.pdf}
   \end{center}
   \caption{
   % \textcolor{red}{
   Visualization of adversarial examples.
   % }
   }
   \label{fig:adv_sample_vis}
\end{figure*}

\captionsetup[subfloat]{font={rm,small},   
                        labelfont={rm,bf}, 
                        textfont={rm}}     
\begin{figure*}[!tb]
    \centering
    \captionsetup[subfloat]{labelformat=parens, listofformat=parens}
    \subfloat[Sim10k $\rightarrow$ Cityscapes]{
        \centering
        \includegraphics[width=0.85\textwidth]{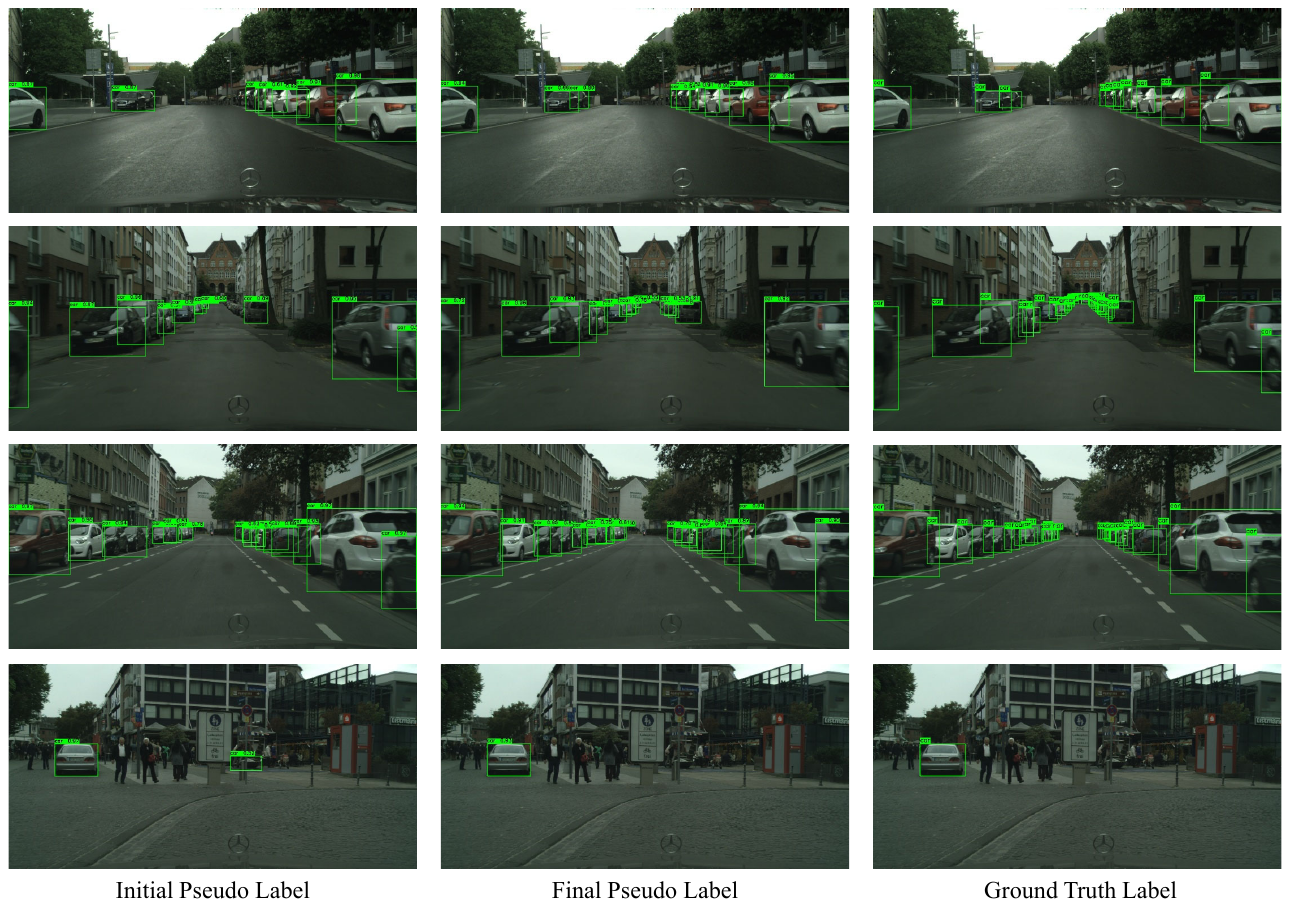}
        \label{fig:ps-vis1}
    }\\[1ex]
    \subfloat[KITTI $\rightarrow$ Cityscapes]{
        \centering
        \includegraphics[width=0.85\textwidth]{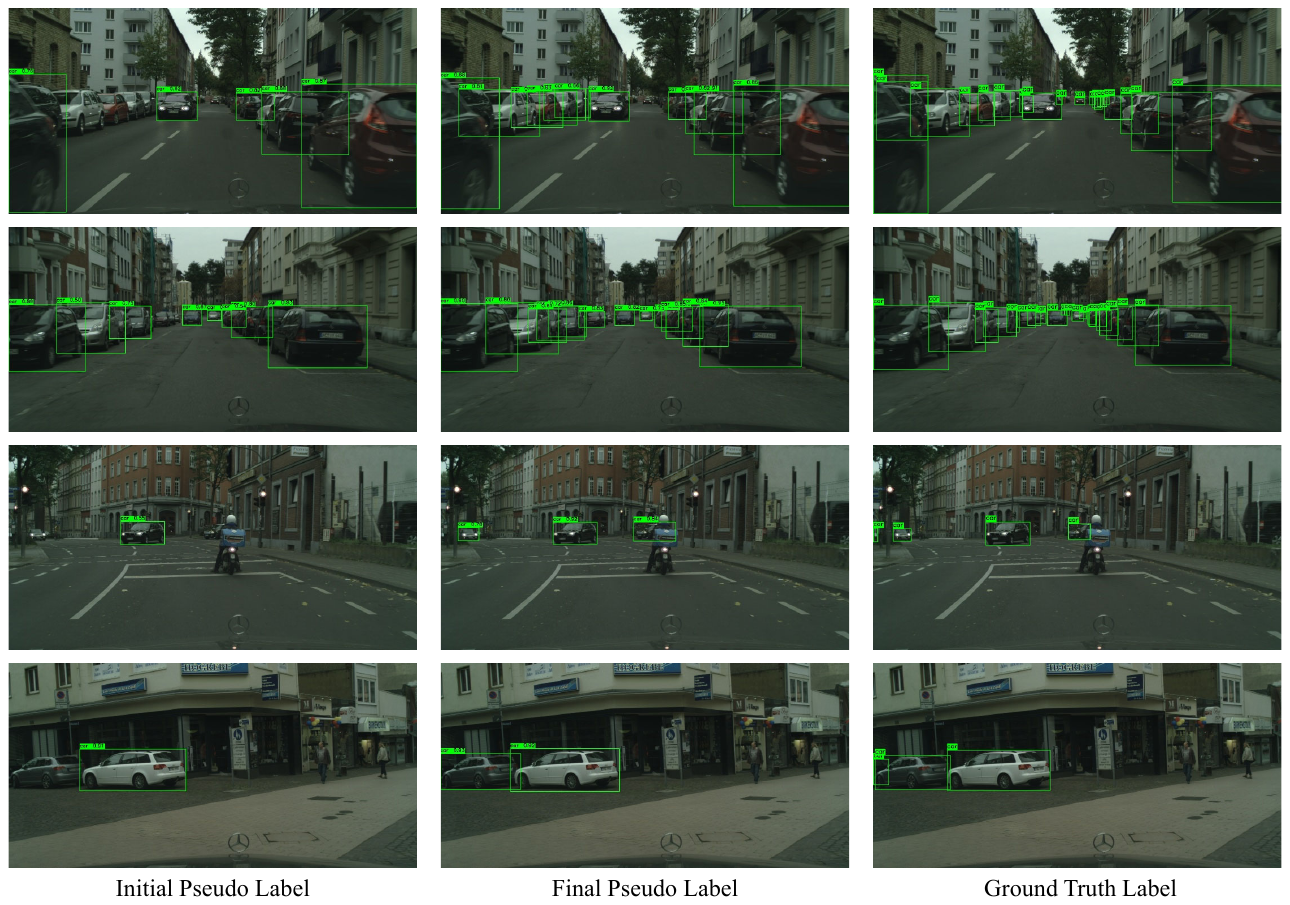}
        \label{fig:ps-vis2}
    }
    \caption{
    % \textcolor{red}{
    Visualization of pseudo labels during training.
    % }
    }
    \label{fig:pseudo_label_evolution}
\end{figure*}

% \clearpage

\begin{figure*}[!tb]\ContinuedFloat
    \centering
    \captionsetup[subfloat]{labelformat=parens, listofformat=parens}
    \subfloat[Pascal VOC $\rightarrow$ Clipart]{
        \centering
        \includegraphics[width=0.47\textwidth]{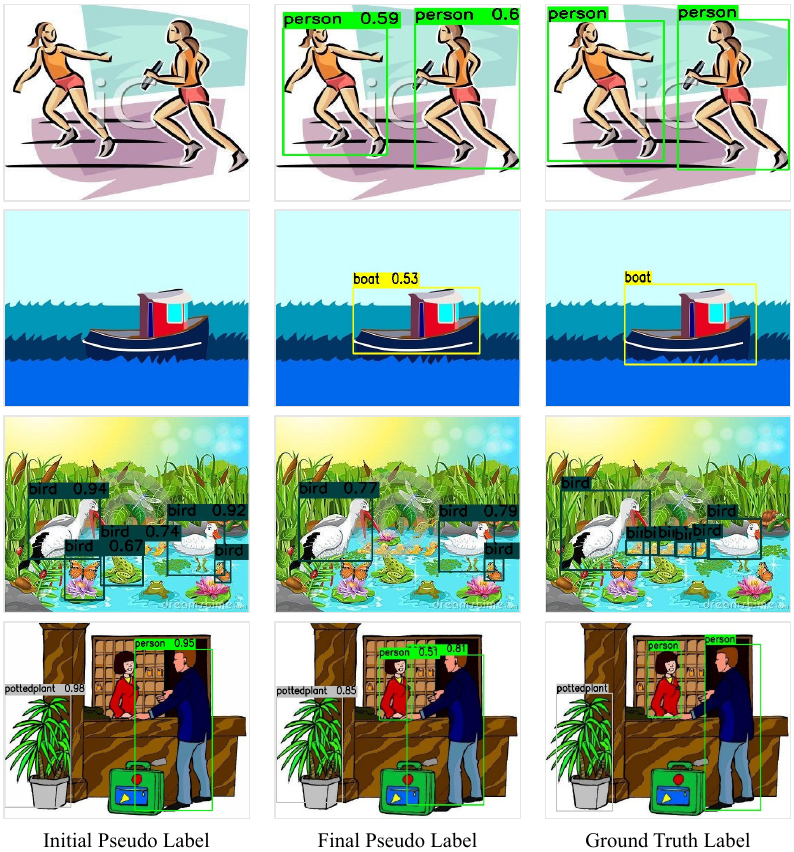}
        \label{fig:ps-vis3}
    }
    \hfill
    \subfloat[Pascal VOC $\rightarrow$ Comic]{
        \centering
        \includegraphics[width=0.47\textwidth]{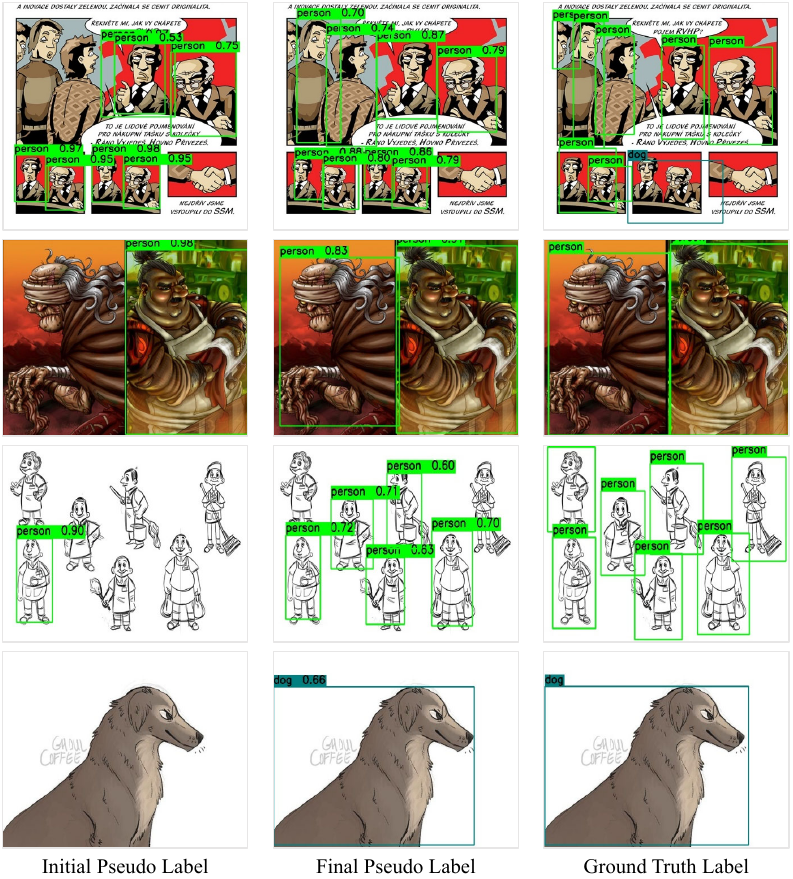}
        \label{fig:ps-vis4}
    }
    \caption{
    % \textcolor{red}{
    Visualization of pseudo labels during training. (cont.)
    % }
    }
\end{figure*}

\textbf{Pseudo-label loss distribution.}
The proposed method in this study effectively addresses the issue of increasing the proportion of simple samples, as depicted in Figure~\ref{fig:pslabel_performance}. 
Updating the model using a previous pseudo-label update strategy~\cite{yang2021st3d} can be challenging due to the large number of simple pseudo labels that arise in later stages of training. 
However, as shown in Figure~\ref{fig:pslabel_performance}, the proposed method can adapt the proportion of simple samples in pseudo labels, thus allowing the model to be updated effectively using the pseudo labels as a supervisory signal.
The experimental results in Table~\ref{tab:ablation} demonstrate that the proposed approach yields a significant improvement of 4.6\% mAP compared with SimROD when the proportion of simple samples is appropriately adjusted.

\textbf{Domain adaptation in various traffic scenarios.}
We conduct additional experiments to analyze the method's performance under different domain shift conditions, using the adaptation setting Sim10k → Cityscapes with the YOLOv5 model. 
Below, we present two experiments that validate DeSimPL’s ability to generalize across significant domain differences in traffic environments.
\textbf{\textit{1) City-specific domain shift.}}
To evaluate DeSimPL’s ability to handle intra-domain shifts across different cities, we compare its performance on the Cityscapes val set (covering multiple cities) and three individual cities: Frankfurt, Lindau, and Munster. The results are summarized in Table~\ref{tab:different_citie}. 
While the overall performance on the entire val set achieves 55.3\% mAP, DeSimPL demonstrates varying performance across cities, with 53.3\% mAP for Frankfurt, 67.6\% mAP for Lindau, and 58.0\% mAP for Munster. 
These results highlight the model's adaptability to different urban environments, with variations reflecting the distinct characteristics of each city, such as traffic density, road layouts, and object appearances.
\textbf{\textit{2) Weather-based domain shift.}}
To assess DeSimPL’s robustness under adverse weather conditions, we evaluate its performance on the Foggy Cityscapes dataset, which simulates varying levels of fog (corresponding to visibility ranges of 600m, 300m, and 150m) and compare the results to the clear-weather Cityscapes val set. 
As shown in Table~\ref{tab:foggy_conditions}, the model achieves 55.3\% mAP in clear weather ($Foggy\_beta=0$) but shows a performance degradation as fog density increased: 52.6\% mAP at $Foggy\_beta=0.005$, 47.9\% mAP at $Foggy\_beta=0.01$, and 40.2\% mAP at $Foggy\_beta=0.02$. 
This demonstrates that while DeSimPL can adapt to mild fog conditions, its performance is increasingly affected by more extreme weather scenarios, which remain a challenging domain shift.

\subsection{\textbf{Visualization and qualitative analysis}}
To further validate the effectiveness and analyze the behavior of the proposed DeSimPL method, we present visualizations organized into three aspects: qualitative comparisons of detection results, adversarial data augmentation examples, and the pseudo-label refinement process.
\par
\textbf{Qualitative comparison of detection results.}
Figure~\ref{fig:result_vis1} and Figure~\ref{fig:result_vis2} compare the detection performance of DeSimPL with the baseline SimROD and alternative methods (D-Adapt, Sigma++) under different adaptation settings. 
We specifically showcase results for both Sim10k $\rightarrow$ Cityscapes adaptation using the YOLOv5 model and Pascal VOC $\rightarrow$ Clipart adaptation. 
The results across these settings highlight DeSimPL's superior ability to detect small, occluded, and distant objects while significantly reducing false positives compared to the baseline. These improvements are particularly evident in complex scenes, emphasizing the method's robustness under challenging domain shifts.

\textbf{Adversarial data augmentation.}
Figure~\ref{fig:adv_sample_vis} presents examples of adversarial images generated using FGSM during the data augmentation step. These visualizations, sampled from all adaptation settings (Sim10k $\rightarrow$ Cityscapes, KITTI $\rightarrow$ Cityscapes, Pascal VOC $\rightarrow$ Clipart, and Pascal VOC $\rightarrow$ Comic), show the subtle perturbations introduced to the input images after DomainMix. This augmentation encourages the model to learn more robust features and increases the proportion of hard samples, contributing to the overall performance improvement by mitigating the simple-label bias.

\textbf{Pseudo-label refinement.}
Figure~\ref{fig:pseudo_label_evolution} demonstrates the dynamic pseudo-label refinement process enabled by our instance-level memory bank and WBF strategy. 
We provide visual examples from all adaptation settings explored in our experiments (Sim10k $\rightarrow$ Cityscapes, KITTI $\rightarrow$ Cityscapes, Pascal VOC $\rightarrow$ Clipart, and Pascal VOC $\rightarrow$ Comic) to illustrate how pseudo labels for target domain objects are progressively improved in terms of both localization accuracy and confidence throughout the adaptation process. 
This visualization confirms the effectiveness of our refinement strategy across diverse scenarios.
\section{Conclusion and Perspectives}
In this work, we have identified a limitation in the self-labeling-based domain adaptive object detection. This limitation arises from an increase in the number of simple samples in pseudo labels as the model trains, leading to a decrease in the gradient update provided by the pseudo label. We have developed DeSimPL, a solution that overcomes this obstacle and improves the effectiveness of self-labeling methods. Our proposed method outperforms domain-alignment methods on multiple benchmark datasets.
\par

While the proposed DeSimPL method achieves strong performance in Domain Adaptive Object Detection (DAOD), certain limitations remain. The method assumes clean source domain annotations, which may not always be available in real-world scenarios. 
Addressing noisy labels~\cite{LiuLYLY22} through robust training strategies could enhance its reliability. Additionally, while extending the method to open-set DAOD~\cite{LiGY23}, where the target domain includes novel object classes, is an interesting direction, this is less critical for transportation tasks with well-defined categories. 
Finally, the reliance on source domain data limits the method’s applicability in scenarios where access to source data is restricted. Developing source-free adaptation techniques~\cite{LiuLY24} is a key priority for future work to ensure broader applicability in real-world intelligent transportation systems.
\par
Furthermore, while our experiments focus on the SimROD framework, the modular design of DeSimPL ensures its generality across different self-labeling methods. The instance-level memory bank, adversarial data augmentation, and adaptive weighted loss are not tied to any specific framework and address fundamental challenges in DAOD. For instance, the memory bank can refine pseudo-label quality in any iterative pseudo-labeling process, while adversarial augmentation and adaptive weighting can enhance robustness and performance in diverse self-labeling paradigms. Future work could explore integrating DeSimPL into other paradigms, such as teacher-student frameworks or ensemble-based methods, to further extend its utility.

\bibliographystyle{IEEEtran}
\bibliography{egbib}

\end{document}